\documentclass{article}

\usepackage{arxiv}

\usepackage[utf8]{inputenc} 
\usepackage[T1]{fontenc}    
\usepackage{hyperref}       
\usepackage{url}            
\usepackage{booktabs}       
\usepackage{amsfonts}       
\usepackage{amsmath}
\usepackage{nicefrac}       
\usepackage{microtype}      
\usepackage{lipsum}		
\usepackage{graphicx}
\usepackage{natbib}
\usepackage{doi}
\usepackage{tcolorbox}
\usepackage{amsmath}
\usepackage{amssymb}
\usepackage{algorithm}
\usepackage{algpseudocode}
\usepackage{caption}
\usepackage{placeins}
\newenvironment{promptbox}[1]{\begin{tcolorbox}[title=#1]}{\end{tcolorbox}}
\usepackage{multirow}

\title{AgenticConflictLens: From Long Traffic Videos to Verifiable Conflict Evidence}

\author{ {\hspace{1mm}Boshra Khalili} \\
	Department of Civil Engineering \\and Engineering Mechanics\\
    Columbia University\\
	New York, NY 10027, USA \\
	\texttt{bk2898@columbia.edu} \\
    \And
    {\hspace{1mm}Sonia Moshfeghi} \\
    	College of Engineering \\ and Computer Science\\
        Florida Atlantic University\\
    	Boca Raton, 33431, FL, USA \\
    	\texttt{smoshfeghi2021@fau.edu} \\
    \And
    {\hspace{1mm}Jinwoo Jang} \\
    	College of Engineering \\ and Computer Science\\
        Florida Atlantic University\\
    	Boca Raton, 33431, FL, USA \\
    	\texttt{jangj@fau.edu}  \\
    \And
    {\hspace{1mm}Andrew Smyth} \\
    	Department of Civil Engineering\\ and Engineering Mechanics\\
        Columbia University\\
    	New York, NY 10027, USA \\
    	\texttt{aws16@columbia.edu} \\
    }

\renewcommand{\shorttitle}{AgenticConflictLens}

\hypersetup{
pdftitle={AgenticConflictLens: From Long Traffic Videos to Verifiable Conflict Evidence},
pdfauthor={Boshra Khalili, Andrew Smyth},
pdfkeywords={Urban Sensing, Intelligent Transportation Systems, Traffic Conflict Detection, Agentic AI, Video Understanding, Long-Video Retrieval},
}

\begin{document}
\maketitle

\begin{abstract}
Traffic cameras continuously record urban road-user behavior, serving as
an important urban sensing resource and providing proactive safety evidence
beyond crash reports. However, safety-critical
events---conflicts, near misses, unsafe turns, sudden braking, and crosswalk
interactions---are brief, rare, and hidden within hours of ordinary traffic,
making large-scale video review costly for transportation agencies. Existing
urban traffic-video analysis pipelines provide detections, tracks, trajectories,
and surrogate-risk scores, but process videos uniformly and do not adapt to the
specific conflict an analyst seeks. Recent vision-language and long-video
reasoning methods improve semantic scene understanding, but often produce
descriptions or answers without traffic-specific verification for localizing
participant-level conflicts and justifying them with motion evidence.
We propose AgenticConflictLens, an agentic, query-conditioned
framework for efficient and auditable conflict extraction from long
fixed-camera urban traffic videos. Given a natural-language safety query,
the system uses a Traffic-Aware Controller, structured traffic memory, and
specialized conflict-analysis tools to decide where to search, which road
users to inspect, what evidence to verify, and which candidates to reject.
A coarse-to-fine process retrieves candidate intervals and applies detailed
verification only where traffic evidence warrants it. The output is a set
of verified conflict records grounded in refined intervals, participant
pairs, semantic traffic regions, trajectories, motion evidence, uncertainty,
rejection reasons, and supporting clips. Evaluated on 96 annotated hours
from Bellevue and NYC DOT, AgenticConflictLens achieves the highest
conflict-existence F1 and temporal IoU on both datasets: 0.84 F1 and
0.77 tIoU on Bellevue, and 0.80 F1 and 0.73 tIoU on NYC DOT, while
inspecting only 8.7\% and 11.9\% of the video, respectively, supporting
scalable, evidence-grounded urban traffic-safety analysis.
\end{abstract}

\providecommand{\method}{}
\renewcommand{\method}{\textsc{AgenticConflictLens}}
\section{Introduction}
\label{sec:introduction}

Traffic cameras provide continuous visual records of road-user behavior at
urban intersections, capturing vehicles, pedestrians, cyclists, buses, trucks,
turning movements, lane changes, occlusions, lighting variation, and weather
effects~\cite{bose2026traffic}. As an important source of urban sensing data,
these cameras offer rich information about traffic interactions and mobility
patterns. Because they observe everyday traffic continuously, these videos
provide proactive safety evidence: they reveal road-user interactions,
conflicts, and near-miss events that may never appear in crash records alone
~\cite{saunier2007automated,staubin2015large,sayed1999traffic}.
However, the safety-critical moments are rare and short. A
vehicle--pedestrian conflict, cyclist--vehicle near miss, sudden braking
event, unsafe turn, risky lane change, or crosswalk-related interaction may
occupy only a few seconds within hours of ordinary traffic
~\cite{arun2021trafficconflict,arun2021systematic,khalili2026streamood,quispe2022iseeyou,
shi2025scvlm}. The central challenge is therefore not simply road-user
detection, but conflict extraction: identifying when the event occurs,
where in the urban scene it happens, which road users are involved, and
what trajectory and motion evidence supports the conflict interpretation
~\cite{singh2025spatiotemporal,arun2022howmany}.

This challenge is important for urban sensing and intelligent transportation
because long traffic-camera videos are valuable but difficult to use at
scale. Transportation agencies and researchers often need to review many
hours of footage to understand where conflicts occur, which road users are
involved, and which locations may require safety intervention
~\cite{staubin2015large,saunier2007automated,wang2026visionlanguage}.
Yet conflict annotations are expensive, require expert review, and cannot
cover every camera view, intersection layout, traffic condition, or safety
question~\cite{arun2021systematic,quispe2022iseeyou,das2023lowcost}.
A practical urban video-analysis system must therefore support targeted
safety investigation: it should search through mostly ordinary traffic,
surface the few intervals that may contain meaningful conflict evidence,
verify whether those intervals truly support a participant-level conflict,
reject weak candidates with explicit reasons, and produce auditable records
that analysts can inspect, compare across sites, and use for downstream
urban traffic-safety planning
~\cite{kuznietsov2024explainable,wang2026visionlanguage}.

Conventional traffic-video analytics usually rely on fixed pipelines:
detect road users, track them, extract trajectories, define conflict zones,
and compute surrogate-risk measures such as Time-to-Collision (TTC),
Post-Encroachment Time (PET), speed, acceleration, headway, and minimum
distance~\cite{arun2021trafficconflict,liu2026video}. Modern trackers
such as ByteTrack~\cite{zhang2022bytetrack} and trajectory-focused
datasets~\cite{quispe2022iseeyou,bock2020ind} make this evidence useful
for urban traffic interaction analysis. However, these pipelines typically
process predefined regions and intervals uniformly. They do not adapt to
the analyst's conflict query, use intermediate verification outcomes to
redirect the search, or decide which participant pairs and intervals warrant
detailed inspection~\cite{razi2023deep}. As a result, they are poorly
suited to long recordings where only a few seconds may contain the conflict
evidence of interest~\cite{ye2025temporal}. This limits their ability to
transform large-scale urban video observations into targeted, interpretable
evidence for traffic-safety analysis.

Recent vision-language models and long-video reasoning methods offer more
flexible semantic scene understanding. Traffic-oriented vision-language
datasets and systems support behavior description, video question answering,
spatiotemporal grounding, and accident understanding in traffic scenes
~\cite{kong2024wts,zhou2025tumtraf,zhang2025seeunsafe,khalili2025autodriveqa, picek2026accident,
marcu2024lingoqa}. Query-adaptive long-video methods further reduce
exhaustive observation through hierarchical search, agentic keyframe selection,
temporal grounding, memory, and tool-guided reasoning
~\cite{wang2025videotree,fan2025akeys,liu2025videomind,
yin2026videoarm,lin2026videoseek,yang2025vca,fan2024videoagent}.
However, these methods are usually designed for description, grounding,
or general VideoQA rather than verified traffic conflict extraction.
They can search selectively, but they do not provide the traffic-specific
verification needed to distinguish a true road-user conflict from ordinary
co-presence, non-closing motion, ambiguous spatial support, or tracking noise
~\cite{li2025vidhalluc,huang2024deconfusetrack,jung2025consistency}.
This gap motivates a traffic-aware control mechanism that combines
query-conditioned long-video search with specialized conflict verification,
integrating visual observations, scene context, and trajectory evidence
for interpretable urban traffic analysis.

We propose \textsc{AgenticConflictLens}, an agentic, query-conditioned
framework for extracting interpretable participant-level conflicts from
long fixed-camera urban traffic videos. The framework is organized around
three connected components: a Traffic-Aware Controller, structured traffic
memory, and specialized conflict-analysis tools. Given a natural-language
query, the controller converts the request into a structured task specifying
the relevant road users, scene regions, interaction type, and evidence
requirements. Scene-prior memory stores reusable fixed-camera urban context,
such as crosswalks, lanes, turning areas, sidewalks, bike lanes, and candidate
conflict regions; evidence and result memory store candidate intervals,
trajectory cues, rejected hypotheses, and verified records. Together, these
components organize visual, spatial, and motion evidence into a structured,
scene-aware representation of candidate traffic interactions. This
organization allows the system to make state-dependent search and
verification decisions instead of running the same fixed analysis over
every interval.

The controller coordinates a coarse-to-fine conflict search. A coarse
scanner first retrieves candidate intervals from long videos by looking
for agreement among traffic-relevant cues, including participant
co-presence, semantic-zone support, proximity, relative motion, path
overlap, and track reliability. These intervals are treated as hypotheses,
not final detections. A fine conflict specialist then verifies each candidate
using trajectory quality, traffic context, visible road-user response,
and uncertainty evidence. Confirmed events become structured conflict
records, while rejected candidates retain explicit rejection reasons.
The workspace is updated after each inspection, so rejections redirect
the search and confirmations become auditable evidence for the final output.

The final output is a set of verified participant-level conflict records
rather than a generic video summary, a single accident label, or an
unstructured answer. Each record is grounded in a refined interval,
participant pairs, semantic traffic region, trajectories, motion evidence,
uncertainty, and a supporting clip. By combining traffic-domain structure,
query-conditioned long-video search, and memory-guided agentic control,
\textsc{AgenticConflictLens} addresses a practical but under-served problem:
extracting few-second conflicts from hours of urban traffic-camera footage
efficiently, accurately, and interpretably. The resulting structured
evidence supports scalable urban sensing and provides actionable information
for traffic-safety assessment and mobility analysis.

The main contributions of this work are as follows:
\begin{enumerate}
\item \textbf{Query-conditioned urban traffic conflict extraction.}
We formulate long traffic-camera conflict extraction as an urban
sensing and safety-review task that recovers rare participant-level
conflicts from hours of video, grounded in time, scene region,
participant pair, and motion evidence.

\item \textbf{Traffic-aware agentic framework.}
We propose \textsc{AgenticConflictLens}, which combines a Traffic-Aware
Controller, structured memory, reusable urban scene priors, and
specialized tools to decide where to search, what to verify, and
when to stop.

\item \textbf{Efficient and auditable real-world evaluation.}
We develop a coarse-to-fine verification process that inspects only
promising intervals, rejects unsupported candidates with explicit
reasons, and produces auditable conflict records for urban traffic-safety
analysis. Because public long-video conflict benchmarks are limited,
we construct annotations and queries for 96 real-world hours from
Bellevue and NYC.
\end{enumerate}

\section{Related Work}
\label{sec:related_work}

Traffic-video conflict analysis is commonly built on fixed perception
pipelines: detecting and tracking road users, extracting trajectories, defining
conflict zones, and computing surrogate-risk measures such as
Time-to-Collision (TTC), Post-Encroachment Time (PET), speed, acceleration,
headway, and minimum distance~\cite{arun2021trafficconflict}. Trackers such as
ByteTrack~\cite{zhang2022bytetrack} provide persistent identities, and datasets
such as I See You~\cite{quispe2022iseeyou} support trajectory-based interaction
analysis. These methods provide interpretable motion evidence, but they usually
process predefined regions and intervals uniformly
~\cite{zhao2021goodpractices}. They do not adapt to a
specific conflict query, use verification outcomes to revise the search, or
decide which participant pair, scene region, interval, or metric deserves
detailed inspection.

Traffic vision-language models extend traffic analysis toward semantic
understanding. WTS~\cite{kong2024wts} supports traffic-behavior description,
TUMTraf VideoQA~\cite{zhou2025tumtraf} studies traffic-oriented question
answering and grounding, and accident-focused methods such as
SeeUnsafe~\cite{zhang2025seeunsafe} and ACCIDENT~\cite{picek2026accident}
address accident classification and localization. These works improve
traffic-scene interpretation, but they are generally organized around
predefined clips, events, or task taxonomies
~\cite{ma2024dolphins}. They do not directly address
query-specified conflict extraction from hours of untrimmed traffic video, where
the output must be grounded in participant identities, precise intervals,
semantic road regions, trajectories, motion evidence, and uncertainty
~\cite{li2025vladbench}.

Long-video reasoning methods reduce exhaustive observation through selective
search, memory, and tool-guided reasoning. VideoTree~\cite{wang2025videotree}
expands query-relevant temporal branches, AKeyS~\cite{fan2025akeys} performs
agent-guided keyframe search, VideoMind~\cite{liu2025videomind} uses
planner--grounder--verifier roles, and VideoARM~\cite{yin2026videoarm} and
VideoSeek~\cite{lin2026videoseek} apply adaptive coarse-to-fine observation.
However, these methods mainly target general VideoQA or temporal grounding,
where evidence is represented as frames, captions, tokens, or broad segments
~\cite{tan2024koala}. Traffic conflict extraction requires
a more structured decision process that can distinguish true conflicts from
ordinary co-presence, non-closing motion, ambiguous spatial support, and
tracking noise
~\cite{zou2026riskprop}.

\textsc{AgenticConflictLens} bridges these directions by combining
query-adaptive long-video search with traffic-specific conflict verification.
It converts a natural-language conflict query into an executable analysis
route, reuses fixed-camera scene context, retrieves candidate intervals through
coarse traffic evidence scanning, and verifies or rejects each candidate with a
fine conflict specialist. The output is an auditable participant-level conflict
record---including participants, refined interval, semantic region, conflict
type, road-user responses, trajectory evidence, uncertainty, and supporting
clip---rather than a generic video summary or accident label.

\section{Method}
\label{sec:method}

\subsection{Overview}
\label{subsec:method_overview}

Given a long fixed-camera traffic video $V=\{I_t\}_{t=1}^{T}$ and a
natural-language conflict query $q$, \textsc{AgenticConflictLens} extracts
query-relevant participant-level road-user conflicts. The output is a set of
verified conflict records
\begin{equation}
\mathcal{C} =
\{(\tau_i, p_i, z_i, k_i, r_i, e_i, u_i, v_i)\}_{i=1}^{N},
\label{eq:conflict_records}
\end{equation}
where $\tau_i=[t_i^{s},t_i^{e}]$ is the refined conflict interval, $p_i$ is the
involved participant pair, $z_i$ is the semantic scene region, $k_i$ is the
conflict type, $r_i$ records the observable road-user responses, $e_i$ stores
trajectory and motion evidence, $u_i$ summarizes uncertainty, and $v_i$ is an
annotated supporting clip. Examples of response signatures include braking,
stopping, yielding, acceleration, and directional adjustment. A coarse flagged
interval is not treated as a final conflict; it is a candidate hypothesis that
must be verified or rejected before record generation.

The framework is organized as a memory-guided traffic-analysis loop rather than
a fixed full-video pipeline. At iteration $\ell$, the \emph{Traffic-Aware
Controller} reads the current traffic workspace $\mathcal{M}_{\ell}$ and
selects the next action
\begin{equation}
a_{\ell}=\pi_{\mathrm{TAC}}(q,\mathcal{M}_{\ell}),
\label{eq:controller_policy}
\end{equation}
where $a_{\ell}$ may invoke a traffic-analysis tool, revise the current search
route, verify a candidate, or stop and generate final records. Each action
writes structured evidence back to memory, producing $\mathcal{M}_{\ell+1}$.
Thus, the executed analysis path depends on the query, scene priors, candidate
quality, verification outcomes, rejected hypotheses, and remaining uncertainty.

\begin{figure*}
    \centering
    \includegraphics[width=0.9\linewidth]{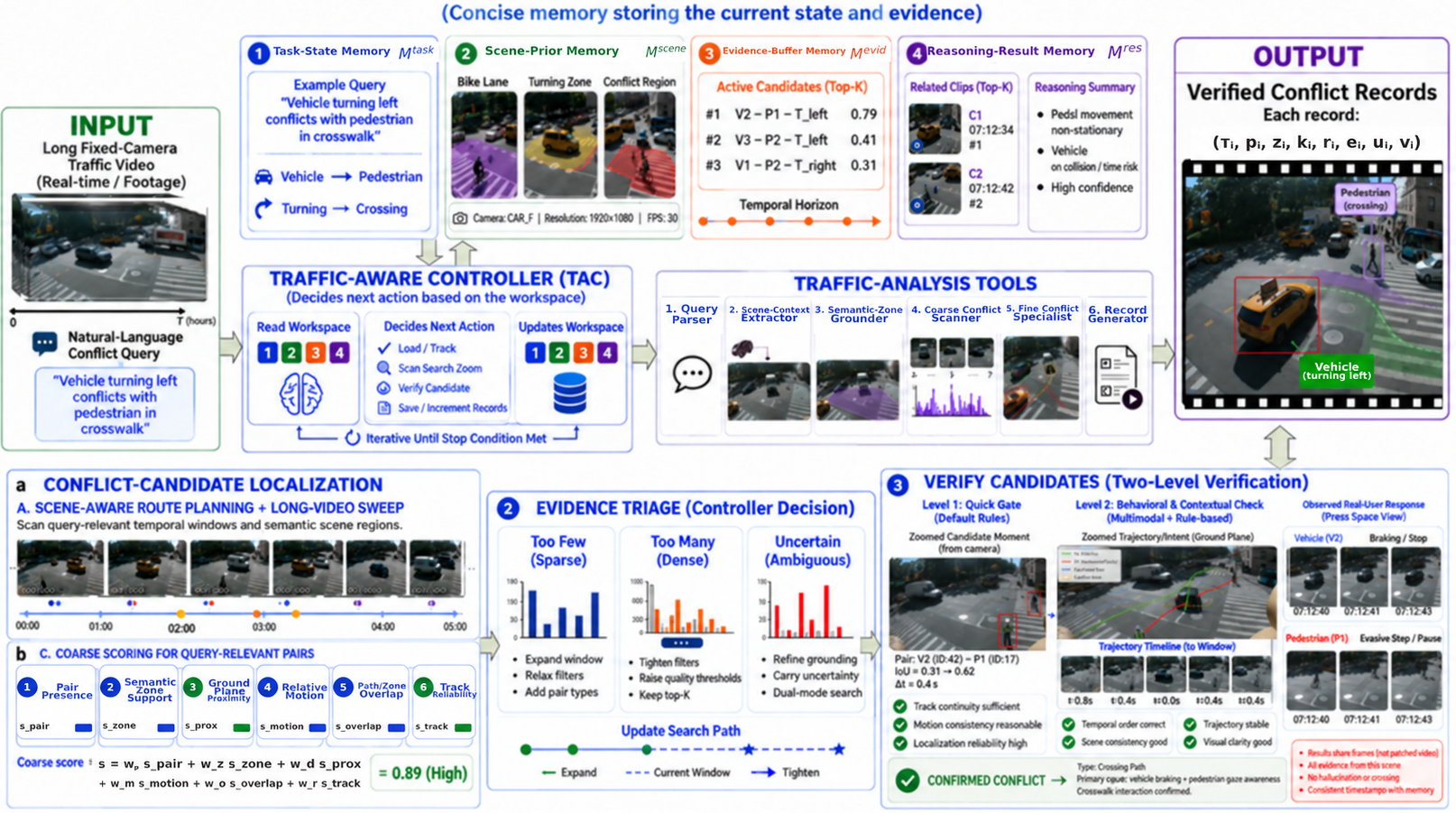}
    \caption{Agentic traffic-conflict analysis pipeline. The Traffic-Aware
    Controller uses structured memory and specialized tools to convert a
    natural-language safety query into a selective coarse-to-fine search,
    verify candidate conflicts, and generate auditable participant-level
    records.}
    \label{fig:agentic_pipeline}
\end{figure*}

\subsection{Online Traffic Workspace}
\label{subsec:workspace}

The Traffic-Aware Controller operates over an online traffic workspace:
\begin{equation}
\mathcal{M}=
\{\mathcal{M}^{task},\mathcal{M}^{scene},
\mathcal{M}^{evid},\mathcal{M}^{res}\}.
\end{equation}
This workspace stores the evolving state of the query-conditioned search,
including query intent, fixed-camera context, inspected intervals, participant
pairs, active candidates, verified conflicts, and rejected hypotheses.

\begin{itemize}
    \item \textbf{Task-state memory.}
    $\mathcal{M}^{task}$ stores the parsed query: temporal and spatial scope,
    participant classes, interaction type, evidence requirements, query
    confidence, and pending action. It determines whether the controller follows
    a focused search route or a broader long-video search.

    \item \textbf{Scene-prior memory.}
    $\mathcal{M}^{scene}$ stores reusable fixed-camera context, such as
    crosswalks, vehicle lanes, sidewalks, bike lanes, curb areas, turning
    regions, and candidate conflict regions. These priors are extracted once
    per camera view and reused to ground trajectory evidence in traffic context.

    \item \textbf{Evidence-buffer memory.}
    $\mathcal{M}^{evid}$ stores query-specific evidence, including selected
    windows, spatial filters, candidate participant pairs, coarse scores,
    sampled visual evidence, trajectory cues, track-quality indicators, and
    flagged intervals. It supports decisions to verify, expand, tighten, or mark
    candidates as uncertain.

    \item \textbf{Reasoning-result memory.}
    $\mathcal{M}^{res}$ stores verified conflicts, rejected candidates, refined
    boundaries, semantic regions, conflict types, road-user responses,
    uncertainty, rejection reasons, and annotated clips. Retaining rejections
    helps the controller avoid repeated weak hypotheses and refine uncertain
    ones.
\end{itemize}

\subsection{Traffic-Analysis Tools}
\label{subsec:traffic_tools}

The Traffic-Aware Controller invokes specialized tools that read the current
workspace, perform one traffic-specific operation, and write structured evidence
back to memory.

\begin{itemize}
    \item \textbf{Query parser.}
    The query parser converts the natural-language query into a structured task
    state, including temporal scope, spatial scope, participant classes,
    interaction type, and evidence requirements. It also determines whether the
    query is interval-focused or requires open-ended long-video search. The
    prompt interface is provided in Appendix 1.2.1.

    \item \textbf{Scene-context extractor.}
    The scene-context extractor builds reusable fixed-camera priors from
    representative frames and camera geometry, including crosswalks, vehicle
    lanes, sidewalks, bike lanes, curb areas, turning regions, and candidate
    conflict regions. It also estimates a fixed-view homography so trajectories
    can be projected to an approximate ground plane for metric distance, speed,
    and TTC estimation. These priors are computed once per camera view and
    reused across queries.

    \item \textbf{Semantic-zone grounder.}
    The semantic-zone grounder maps query regions, such as ``near the
    crosswalk'' or ``in the turning lane,'' to scene-prior memory and writes
    grounded zone constraints and spatial uncertainty back to the workspace.

    \item \textbf{Coarse conflict scanner.}
    The coarse conflict scanner performs first-pass localization over the
    planned windows, semantic regions, and participant-pair types selected by
    the controller. It returns ranked flagged intervals with coarse motion,
    trajectory, zone, and track-quality evidence.

    \item \textbf{Fine conflict specialist.}
    The fine conflict specialist verifies selected flagged intervals by
    combining track quality, context-adaptive ground-plane proximity, closing or
    converging motion, semantic-zone consistency, trajectory evidence, TTC
    evidence when available, and visible road-user response.

    \item \textbf{Record generator.}
    The record generator reads verified and rejected outcomes from
    $\mathcal{M}^{res}$ and produces the final auditable output: verified
    conflict records, annotated clips, uncertainty notes, and rejection
    summaries. The prompt interface is provided in Appendix 1.2.5.
\end{itemize}

\subsection{Traffic-Aware Controller-Driven Pipeline}
\label{subsec:controller_driven_pipeline}

\textsc{AgenticConflictLens} follows a traffic-specific control loop:
\emph{sense}, \emph{scene-aware route planning},
\emph{conflict-candidate localization}, \emph{evidence triage},
\emph{verify}, and \emph{record and memory update}. The stages define the
typical workflow, but the Traffic-Aware Controller can skip, repeat, expand,
tighten, or stop them based on the current workspace.

\begin{itemize}
    \item \textbf{Sense.}
    The controller parses the query and writes the task state to
    $\mathcal{M}^{task}$. It then checks whether scene priors exist for the
    camera view; if not, it invokes the scene-context extractor. If the query
    mentions or implies a traffic region, such as a crosswalk, bike lane, curb,
    vehicle lane, or turning area, the controller invokes the semantic-zone
    grounder. This initializes the workspace with query intent, relevant scene
    regions, participant constraints, and evidence requirements.

    \item \textbf{Scene-aware route planning.}
    Using task-state and scene-prior memory, the controller builds an analysis
    route specifying which temporal windows, semantic regions, road-user
    classes, participant-pair types, and evidence requirements should guide the
    search. Focused queries create narrow routes over specified times or
    regions, while open-ended queries create broader routes over larger
    candidate sets. The route is written to $\mathcal{M}^{evid}$ and used for
    candidate localization. The route-planning prompt is provided in
    Appendix 1.2.2.

    \item \textbf{Conflict-candidate localization.}
    The controller invokes the coarse conflict scanner over the planned route.
    This stage reduces long video to a compact set of candidate intervals for
    fine verification. The scanner uses reusable detections, tracks, sampled
    frames, semantic-zone priors, participant constraints, trajectories, and
    query-derived filters, and evaluates only query-relevant participant pairs
    rather than all road users at all times.

    For each temporal bin $b$ and candidate pair $(a_m,a_n)$, the scanner
    computes
    \begin{equation}
    \begin{aligned}
    s(b,a_m,a_n) ={}&
    w_p s_{\mathrm{pair}} + w_z s_{\mathrm{zone}}
    + w_d s_{\mathrm{prox}} \\
    &+ w_m s_{\mathrm{motion}} + w_o s_{\mathrm{overlap}}
    + w_r s_{\mathrm{track}} .
    \end{aligned}
    \label{eq:coarse_score_main}
    \end{equation}
    The terms measure pair presence, semantic-zone support, ground-plane
    proximity, relative or closing motion, path or zone overlap, and track
    reliability. This score ranks candidate evidence only; it does not assign
    final conflict labels. High-scoring bins are merged into flagged intervals
    and stored with participant pair, semantic-region support, visual evidence,
    trajectory cues, coarse score, and track-quality information. Score-term
    details are provided in Appendix 1.1.

    \item \textbf{Evidence triage.}
    Before verification, the controller reviews $\mathcal{M}^{evid}$ because
    coarse search can miss conflicts when filters are too strict, or return many
    weak co-presence candidates in dense traffic. If candidates are too sparse,
    it expands nearby windows, revisits scene grounding, or adds
    query-compatible participant pairs. If candidates are too broad, it
    prioritizes intervals with strong query match, reliable tracks, clear
    semantic-zone support, and closing-motion evidence. Ambiguous candidates are
    refined or carried forward with uncertainty. Additional route-revision
    details are provided in Appendices 1.1.2 and 1.2.3.

    \item \textbf{Verify.}
    For each selected flagged interval, the controller invokes the fine conflict
    specialist to decide whether the candidate is a true participant-level
    conflict or only ordinary co-presence, weak interaction, spatial ambiguity,
    or tracking noise. The specialist reads the candidate pair, grounded
    semantic region, tracks, trajectory cues, sampled visual evidence, coarse
    score, and scene priors from memory. Unlike the coarse scanner, which ranks
    plausibility, the fine specialist makes the final conflict decision.

    The specialist first rejects candidates with severe identity switches,
    implausible jumps, insufficient common observation, or unreliable
    localization. Surviving candidates are verified using minimum ground-plane
    separation, context-adaptive proximity, closing or converging motion,
    semantic-zone consistency, trajectory overlap, TTC evidence when available,
    and visible road-user response. TTC is used only as corroborating evidence
    for closing pairs. A candidate is confirmed only when distance, motion,
    scene-region, trajectory, and response evidence jointly support a conflict.

    Confirmed candidates are assigned a kinematic conflict type, such as
    yielding, turning-path, crossing-path, merging--following, or
    close-proximity. The specialist also extracts response signatures, refines
    the interval to the approach--response--separation span, and records
    uncertainty from separation margin, TTC agreement, track reliability,
    semantic-zone confidence, and visual support. Rejected candidates are
    retained with explicit reasons, such as ordinary co-presence, non-closing
    motion, excessive separation, insufficient track quality, ambiguous spatial
    support, or lack of visible response. Additional localization equations,
    typology rules, response signatures, adaptive-threshold checks, and
    hotspot-zone profiles are provided in Appendices 1.1 and 2.6.2.

    \item \textbf{Record and memory update.}
    After verification, the controller updates $\mathcal{M}^{res}$ with
    confirmed records, rejected candidates, refined intervals, uncertainty, and
    rejection reasons. If the query is satisfied, it invokes the record
    generator; otherwise, it revises the search with updated windows, filters,
    or participant-pair hypotheses. If revisions return only rejected or
    low-confidence candidates, the controller reports a negative or uncertain
    result with searched windows, candidate counts, rejection reasons, and
    remaining uncertainty. The memory-update prompt and stopping rules are
    provided in Appendices 1.2.4 and 1.1.5.
\end{itemize}

\subsection{Final Records}
\label{subsec:final_records}

The record generator produces auditable participant-level outputs linking each
verified conflict to its refined interval, participant pair, semantic region,
conflict type, responses, trajectory evidence, uncertainty, and annotated clip.
Rejected flags are retained with reasons, showing which plausible events were
inspected and dismissed.


\section{Experiments}
\label{sec:experiments}
We evaluate \method{} on 96 annotated hours of real-world fixed-camera traffic
video from Bellevue and NYC DOT. The experiments test query-driven conflict
localization, participant-pair identification, traffic-specific verification,
and search efficiency. We compare against adapted long-video reasoning and
traffic-safety baselines, and ablate the controller, memory, and tools. 
Full dataset, protocol, metric, ablation, and spatial-risk
analysis details are provided in Appendix~2.

\subsection{Datasets}
\label{sec:datasets}

Public long-video traffic benchmarks with participant-level conflict
annotations, natural-language queries, temporal boundaries, semantic regions,
and observable response labels remain limited. We therefore construct a
query-conditioned conflict-extraction evaluation from real-world fixed-camera
videos in two cities.

We evaluate \method{} on 96 annotated hours of real-world
fixed-camera traffic video: 60 hours from five public
Bellevue intersections~\cite{bellevue_traffic_dataset}
and 36 hours from three NYC DOT intersections.
Each site contributes twelve one-hour recordings. For
each video, we construct participant-level conflict annotations and
natural-language evaluation queries with temporal boundaries, participant
pairs, semantic scene regions, conflict types, and observable road-user
responses.

The two datasets provide complementary evaluation settings. Bellevue contains
sparse conflicts separated by long periods of ordinary traffic, testing whether
the system can reduce hours of video to short candidate intervals without
missing rare events. NYC DOT contains denser traffic and more frequent plausible
interactions, testing whether the system can select the requested conflict from
many coarse candidates and reject ordinary co-presence. Together, the videos
cover diverse viewpoints, intersection layouts, road-user mixes, traffic
densities, lighting conditions, occlusions, and small road users.

The Bellevue traffic videos are publicly available
through the City of Bellevue Traffic Video Dataset
~\cite{bellevue_traffic_dataset}, while the NYC DOT
videos are restricted operational recordings.

\paragraph{Annotations and queries.}
We create annotations and queries to evaluate the full output of \method{},
not just clip-level classification. A conflict is a temporally localized
encounter where road users show converging, competing, or mutually dependent
movement with a visible response, such as yielding, braking, stopping,
directional adjustment, gap acceptance, or evasive motion. Proximity alone is
not sufficient. Each conflict is annotated with the method's output schema:
refined interval, participant pair, semantic scene region, conflict type,
primary response, and evidence statement.

The real-world evaluation contains 1,127 annotated conflicts and 4,582 queries
across 96 hours: 119 conflicts and 457 queries from Bellevue, and 1,008
conflicts and 4,125 queries from NYC DOT. Queries are generated from verified
conflict records and manually checked for answerability, temporal scope,
participant constraints, and consistency with the reference annotation. They
cover focused retrieval, open-ended long-video search, participant-pair search,
temporal localization, interaction reasoning, and response reasoning. All main
evaluation queries correspond to annotated conflict events. Additional details
are provided in Appendix 2.1.

\subsection{Evaluation Protocol and Implementation}
\label{sec:protocol_implementation}

\begin{itemize}
    \item \textbf{Protocol.}
    To the best of our knowledge, existing long-video reasoning and traffic-safety
    methods do not directly produce query-conditioned participant-level conflict
    records from long traffic-camera videos. We therefore evaluate the closest
    related methods under a shared conflict-localization interface. Each baseline
    keeps its core search idea while using the same conflict query and returning
    candidate intervals and participant pairs. This enables fair comparison on
    shared outputs, while full structured-record evaluation is reported for
    \method{} and its ablations.
    \item \textbf{Baselines.}
    We evaluate five adapted baselines. VideoTree-inspired retrieval
    ~\cite{wang2025videotree} tests hierarchical coarse-to-fine search and
    returns top retrieved conflict clips. SeeUnsafe-inspired search
    ~\cite{zhang2025seeunsafe} tests direct accident-aware VLM judgment and
    returns clips judged unsafe or conflict-relevant by the VLM.
    VideoARM-inspired retrieval~\cite{yin2026videoarm} tests long-video memory
    and returns memory-selected conflict clips. VideoSeek-inspired search
    ~\cite{lin2026videoseek} tests tool-guided seeking and returns
    tool-selected conflict clips. SSAM-inspired search~\cite{ssam2008manual}
    tests trajectory-based surrogate-safety analysis and returns high-risk
    trajectory intervals. In contrast, \method{} returns verified refined
    conflict intervals and writes them as structured participant-level conflict
    records. All VLM-based methods use the same frozen Qwen2.5-VL-7B-Instruct
    backbone without fine-tuning. Full baseline implementation details are provided in Appendix 2.2.

    \item \textbf{Metrics.}
    All methods are compared using the metrics reported in
    Table~\ref{tab:main_results}: conflict-existence F1, mean
    temporal IoU, participant-pair accuracy, inspection percentage, and runtime.
    Exist. F1 measures event-level conflict retrieval.
    Mean tIoU measures temporal localization quality for matched predictions.
    Pair Acc. measures whether the matched prediction identifies the correct
    participant pair. Inspect \% is the percentage of the query video examined
    by the method, and runtime is reported in minutes per query.
    Structured-record accuracy is reported only for \method{} and its
    ablations, where the full record schema is available. Detailed metric
    definitions and post-processing rules are provided in Appendix 2.2.
\end{itemize}

\subsection{Main Results}
\label{sec:main_results}

Table~\ref{tab:main_results} reports query-driven conflict-localization results
on Bellevue and NYC DOT. \method{} achieves the best overall trade-off between
accuracy and efficiency: it obtains the highest Exist. F1 and tIoU on both
datasets while inspecting the smallest fraction of video. SSAM-style trajectory
scoring has strong pair accuracy because it uses dense motion evidence, but it
requires full-video inspection and higher runtime. Uniform clip search also
inspects much of the video, while selective long-video baselines reduce cost but
remain less accurate than \method{}. Qualitative results, including 
a successful case and a failure trace, are provided in Appendix 2.3.
\begin{table*}[t]
\centering
\scriptsize
\setlength{\tabcolsep}{3.2pt}
\caption{Main query-driven participant-level conflict-localization results on
Bellevue (60 annotated hours) and NYC DOT (36 annotated hours). Each query is
evaluated over a 30-minute video. Avg. Runtime reports the average minutes
required to answer one query over that 30-minute video. Inspect \% is the
percentage of video searched or sent for inspection. Arrows indicate the
preferred direction.}
\label{tab:main_results}
\resizebox{\textwidth}{!}{
\begin{tabular}{lccccc ccccc}
\toprule
&
\multicolumn{5}{c}{\textbf{Bellevue}} &
\multicolumn{5}{c}{\textbf{NYC DOT}} \\
\cmidrule(lr){2-6}
\cmidrule(lr){7-11}
\textbf{Method}
& \textbf{Exist. F1 $\uparrow$}
& \textbf{tIoU $\uparrow$}
& \textbf{Pair Acc. $\uparrow$}
& \textbf{Inspect \% $\downarrow$}
& \textbf{Avg. Runtime $\downarrow$}
& \textbf{Exist. F1 $\uparrow$}
& \textbf{tIoU $\uparrow$}
& \textbf{Pair Acc. $\uparrow$}
& \textbf{Inspect \% $\downarrow$}
& \textbf{Avg. Runtime $\downarrow$} \\
\midrule

High-risk trajectory intervals~\cite{ssam2008manual}
& 0.74 & 0.66 & \textbf{0.86} & 100.0 & 9.8
& 0.71 & 0.64 & 0.80 & 100.0 & 10.5 \\

Uniform sampled clip conflict search~\cite{zhang2025seeunsafe}
& 0.59 & 0.45 & 0.60 & 62.2 & 6.7
& 0.60 & 0.44 & 0.61 & 59.7 & 7.9 \\

Top retrieved conflict clips~\cite{wang2025videotree}
& 0.62 & 0.57 & 0.64 & 23.5 & 3.8
& 0.57 & 0.59 & 0.69 & 29.4 & 4.6 \\

Memory-selected conflict clips~\cite{yin2026videoarm}
& 0.76 & 0.63 & 0.63 & 23.0 & 3.5
& 0.70 & 0.61 & 0.62 & 25.9 & 4.2 \\

Tool-selected conflict clips~\cite{lin2026videoseek}
& 0.73 & 0.71 & 0.75 & 19.1 & 3.1
& 0.74 & 0.63 & 0.72 & 22.7 & 3.8 \\

\method{}
& \textbf{0.84} & \textbf{0.77} & 0.85 & \textbf{8.7} & \textbf{1.6}
& \textbf{0.80} & \textbf{0.73} & \textbf{0.81} & \textbf{11.9} & \textbf{2.1} \\

\bottomrule
\end{tabular}
}
\end{table*}

\subsection{Ablation Study}
\label{sec:ablation_study}

We ablate the main components of \method{}: traffic-analysis tools, structured
traffic memory, and the Traffic-Aware Controller. All variants use the same
videos, queries, backbone, output schema, prompts, and inference budget unless
the ablated component changes the budget. Unlike the adapted baselines in
Sec.~\ref{sec:protocol_implementation}, all ablations remain variants of
\method{} and produce structured participant-level conflict records. We report
shared localization metrics and Record Acc., which measures whether the final
record has the correct interval, participant pair, semantic region, conflict
type, and primary response. Additional details are provided in
Appendix 2.6.

\subsubsection{Traffic-Analysis Tools}
\label{sec:ablation_tools}

Table~\ref{tab:tool_ablation_main} reports the tool ablation results. Each row
removes one tool while keeping the rest of the system unchanged.

\begin{table}[t]
\centering
\scriptsize
\setlength{\tabcolsep}{4pt}
\caption{Ablation on traffic-analysis tools for Bellevue
and NYC DOT. QZ: query parsing and zone grounding; SC:
scene context; CS: traffic-aware coarse scanner; FS: fine conflict specialist.
Record Acc. evaluates the full structured conflict record. Arrows indicate the
preferred direction.}
\label{tab:tool_ablation_main}
\begin{tabular}{cccc cccc}
\toprule
\multicolumn{4}{c}{\textbf{Traffic-Analysis Tools}} &
\multicolumn{4}{c}{\textbf{Metrics}} \\
\cmidrule(lr){1-4}\cmidrule(lr){5-8}
\textbf{QZ} &
\textbf{SC} &
\textbf{CS} &
\textbf{FS} &
\textbf{Exist. F1 $\uparrow$} &
\textbf{tIoU $\uparrow$} &
\textbf{Pair Acc. $\uparrow$} &
\textbf{Record Acc. $\uparrow$} \\
\midrule

\multicolumn{8}{l}{\textbf{Bellevue}} \\

 & \checkmark & \checkmark & \checkmark
& 0.72 & 0.64 & 0.72 & 0.64 \\

\checkmark &  & \checkmark & \checkmark
& 0.79 & 0.73 & 0.79 & 0.7 \\

\checkmark & \checkmark &  & \checkmark
& 0.67 & 0.60 & 0.77 & 0.61 \\

\checkmark & \checkmark & \checkmark &
& 0.73 & 0.62 & 0.82 & 0.59 \\

\checkmark & \checkmark & \checkmark & \checkmark
& \textbf{0.84} & \textbf{0.77} & \textbf{0.85} & \textbf{0.78} \\

\midrule
\multicolumn{8}{l}{\textbf{NYC DOT}} \\

 & \checkmark & \checkmark & \checkmark
& 0.67 & 0.61 & 0.69 & 0.58 \\

\checkmark &  & \checkmark & \checkmark
& 0.75 & 0.67 & 0.74 & 0.62 \\

\checkmark & \checkmark &  & \checkmark
& 0.62 & 0.59 & 0.72 & 0.56 \\

\checkmark & \checkmark & \checkmark &
& 0.68 & 0.59 & 0.79 & 0.56 \\

\checkmark & \checkmark & \checkmark & \checkmark
& \textbf{0.80} & \textbf{0.73} & \textbf{0.81} & \textbf{0.73} \\

\bottomrule
\end{tabular}
\end{table}

The largest drop in Exist. F1 occurs when the coarse scanner is removed,
showing that candidate localization is the main driver of conflict retrieval.
Removing the fine specialist produces the lowest Record Acc., indicating that
coarse flags alone are not sufficient for reliable final records. The full
configuration performs best across both datasets, confirming that search,
scene grounding, query grounding, and verification are complementary.

\subsubsection{Structured Traffic Memory}
\label{sec:ablation_memory}

Table~\ref{tab:memory_ablation_main} evaluates how structured memory affects
long-video conflict search. The context-only variant removes explicit memory and
relies on ordinary in-context reasoning.

\begin{table}[t]
\centering
\scriptsize
\setlength{\tabcolsep}{4pt}
\caption{Ablation on structured traffic memory for Bellevue and NYC DOT. TS:
task-state memory; SP: scene-prior memory; EB: evidence-buffer memory; RR:
reasoning-result memory.}
\label{tab:memory_ablation_main}
\begin{tabular}{ccccc cccc}
\toprule
\multicolumn{5}{c}{\textbf{Memory Configuration}} &
\multicolumn{4}{c}{\textbf{Metrics}} \\
\cmidrule(lr){1-5}\cmidrule(lr){6-9}
\textbf{TS} &
\textbf{SP} &
\textbf{EB} &
\textbf{RR} &
\textbf{Context Only} &
\textbf{Exist. F1} &
\textbf{tIoU} &
\textbf{Pair Acc.} &
\textbf{Record Acc.} \\
\midrule

\multicolumn{9}{l}{\textbf{Bellevue}} \\

 & \checkmark & \checkmark & \checkmark &
& 0.76 & 0.68 & 0.78 & 0.67 \\

\checkmark &  & \checkmark & \checkmark &
& 0.77 & 0.67 & 0.80 & 0.66 \\

\checkmark & \checkmark &  & \checkmark &
& 0.70 & 0.64 & 0.76 & 0.62 \\

\checkmark & \checkmark & \checkmark &  &
& 0.82 & 0.73 & 0.83 & 0.73 \\

 &  &  &  & \checkmark
& 0.69 & 0.60 & 0.71 & 0.56 \\

\midrule
\checkmark & \checkmark & \checkmark & \checkmark &
& \textbf{0.84} & \textbf{0.77} & \textbf{0.85} & \textbf{0.78} \\

\midrule
\multicolumn{9}{l}{\textbf{NYC DOT}} \\

 & \checkmark & \checkmark & \checkmark &
& 0.71 & 0.63 & 0.73 & 0.61 \\

\checkmark &  & \checkmark & \checkmark &
& 0.72 & 0.62 & 0.75 & 0.60 \\

\checkmark & \checkmark &  & \checkmark &
& 0.66 & 0.58 & 0.70 & 0.56 \\

\checkmark & \checkmark & \checkmark &  &
& 0.77 & 0.66 & 0.76 & 0.67 \\

 &  &  &  & \checkmark
& 0.63 & 0.54 & 0.66 & 0.51 \\

\midrule
\checkmark & \checkmark & \checkmark & \checkmark &
& \textbf{0.80} & \textbf{0.73} & \textbf{0.81} & \textbf{0.73} \\

\bottomrule
\end{tabular}
\end{table}

The results show that explicit memory is important on both datasets. The
context-only variant performs worst, indicating that long-video conflict search
requires persistent search state and evidence history. Removing
evidence-buffer memory causes the largest drop, showing that carrying candidate
intervals, tracks, and intermediate evidence from coarse search to fine
verification is critical. Task-state and scene-prior memory support query
consistency and localization, while reasoning-result memory helps preserve
verified and rejected hypotheses. The full memory configuration achieves the
best overall performance.

\subsubsection{Traffic-Aware Controller}
\label{sec:ablation_controller}

Table~\ref{tab:controller_ablation_main} evaluates whether the controller
improves conflict extraction beyond simply making the tools and memory
available. All variants keep the same inputs, tools, memory modules, prompts,
backbone, and output schema, and only remove one controller policy.

\begin{table}[t]
\centering
\scriptsize
\setlength{\tabcolsep}{3.5pt}
\caption{Ablation on Traffic-Aware Controller policies for Bellevue and NYC DOT.
RP: scene-aware route planning; C2F: adaptive coarse-to-fine candidate
promotion; FB/S: verification feedback and adaptive stopping.}
\label{tab:controller_ablation_main}
\resizebox{\linewidth}{!}{
\begin{tabular}{ccc ccccc}
\toprule
\multicolumn{3}{c}{\textbf{Controller Configuration}} &
\multicolumn{5}{c}{\textbf{Metrics}} \\
\cmidrule(lr){1-3}\cmidrule(lr){4-8}
\textbf{RP} &
\textbf{C2F} &
\textbf{FB/S} &
\textbf{Exist. F1 $\uparrow$} &
\textbf{tIoU $\uparrow$} &
\textbf{Record Acc. $\uparrow$} &
\textbf{Reject \% } &
\textbf{Inspect \% $\downarrow$} \\
\midrule

\multicolumn{8}{l}{\textbf{Bellevue}} \\

 & \checkmark & \checkmark
& 0.73 & 0.65 & 0.64 & 73.2 & 19.6 \\

\checkmark &  & \checkmark
& 0.76 & 0.66 & 0.66 & 64.4 & 13.9 \\

\checkmark & \checkmark &
& 0.81 & 0.72 & 0.73 & 68.1 & 12.5 \\

\midrule
\checkmark & \checkmark & \checkmark
& \textbf{0.84} & \textbf{0.77} & \textbf{0.78}
& \textbf{61.7} & \textbf{8.7} \\

\midrule
\multicolumn{8}{l}{\textbf{NYC DOT}} \\

 & \checkmark & \checkmark
& 0.70 & 0.61 & 0.57 & 77.3 & 25.2 \\

\checkmark &  & \checkmark
& 0.72 & 0.62 & 0.61 & 66.8 & 18.7 \\

\checkmark & \checkmark &
& 0.75 & 0.65 & 0.66 & 71.2 & 15.1 \\

\midrule
\checkmark & \checkmark & \checkmark
& \textbf{0.80} & \textbf{0.73} & \textbf{0.73}
& \textbf{63.5} & \textbf{11.9} \\

\bottomrule
\end{tabular}
}
\end{table}

The strongest effect comes from route planning: without it, the system inspects
substantially more video and loses accuracy, showing that the controller must
search in the right scene regions and time windows. Removing adaptive
coarse-to-fine promotion also reduces performance, indicating that fixed
candidate selection is less effective than evidence-aware triage. Removing
feedback and stopping has a smaller effect on retrieval but lowers record
accuracy and increases inspection, showing that verification outcomes help
refine the search and avoid unnecessary processing. The full controller gives
the best accuracy-efficiency trade-off on both datasets.

\FloatBarrier
\section{Limitations}
\label{sec:limitations}

\textsc{AgenticConflictLens} is designed for fixed-camera traffic videos and
depends on reliable scene priors, detections, tracks, and trajectory evidence.
Its controller can store uncertainty, retain rejection reasons, and redirect the
search, but it cannot recover conflicts that are not visually observable or are
lost because of severe occlusion, missed detections, identity switches, poor
lighting, or ambiguous scene geometry.

\section{Conclusion}
\label{sec:conclusion}

We presented AgenticConflictLens, a query-conditioned framework for
extracting rare participant-level conflicts from long fixed-camera traffic
videos. The method uses a Traffic-Aware Controller, structured traffic memory,
and specialized tools to perform selective coarse-to-fine verification rather
than uniform full-video processing. It produces auditable conflict records
grounded in refined intervals, participant pairs, semantic regions, motion
evidence, uncertainty, rejection reasons, and supporting clips.

On 96 annotated hours from Bellevue and NYC DOT, AgenticConflictLens
achieves the highest conflict-existence F1 and temporal IoU on both datasets:
0.84 F1 and 0.77 tIoU on Bellevue, and 0.80 F1 and 0.73 tIoU on NYC DOT, while
inspecting only 8.7\% and 11.9\% of the video, respectively.
\section{Acknowledgments}
This work was supported by the Center for Smart Streetscapes, a National Science Foundation (NSF) Engineering Research Center, under Grant No. EEC-2133516. The author also gratefully acknowledges support from the NYMTC/CAIT September 11th Memorial Program for Regional Transportation Planning Fellowship, as well as the fellowship program and New York City Department of Transportation (NYC DOT) teams for their valuable support and collaboration.

\bibliographystyle{unsrtnat}
\bibliography{references}  


\clearpage
\section*{Appendix}

\setcounter{section}{0}
\setcounter{subsection}{0}
\setcounter{subsubsection}{0}

\section{Additional Method Details}
\label{app:method_details}

This appendix provides implementation details that support
Section 3. Rather than repeating the full Traffic-Aware
Controller pipeline, we specify the evidence used by the coarse conflict
scanner, the route-revision rules used during evidence triage, and the
trajectory-level verification performed by the fine conflict specialist. We also
describe the automatic conflict typology, kinematic response signatures,
adaptive stopping behavior, and prompt interfaces used by the controller.

All geometric evidence is computed from smoothed trajectories projected to an
approximate ground plane. For each fixed camera view, we estimate a homography
from scene reference points and known road-width cues, mapping image-plane
tracks to ground-plane trajectories. This enables distance, speed, TTC, response
signature, and surrogate-risk estimates within each fixed view. Final conflict
decisions are not made from geometry alone; they require agreement among
trajectory, motion, scene-region, visual-response, and uncertainty evidence. The
language-dependent prompt interfaces used by the controller are provided in
Appendix~\ref{app:prompt_details}.

\subsection{Conflict-Candidate Localization and Verification Details}
\label{app:localization_details}

This section provides additional implementation details for the coarse
conflict-candidate localization stage, evidence triage, fine verification,
conflict typing, response signatures, and adaptive stopping. The coarse stage is
used only to retrieve promising intervals. Final conflict decisions are made by
the fine conflict specialist using trajectory, motion, scene-region, response,
and uncertainty evidence.

\subsubsection{Conflict-Candidate Localization Score Terms}
\label{app:coarse_score_terms}

The conflict-candidate localization stage implements the coarse retrieval step
of Section 3.4. For each temporal bin $b$
and candidate participant pair $(a_m,a_n)$ selected by the Traffic-Aware
Controller, the coarse scanner computes the score in
Eq 4 in main text. The score ranks candidate evidence within
query-selected temporal windows, semantic regions, and participant-pair types.
It is used only for retrieval; final conflict decisions are made later by the
fine conflict specialist.

\paragraph{Pair presence.}
$s_{\mathrm{pair}}$ measures whether the query-relevant participant classes
co-occur in the temporal bin with sufficient temporal overlap. For example, a
vehicle--pedestrian query suppresses vehicle--vehicle and
pedestrian--pedestrian pairs unless the query requests general conflict search.

\paragraph{Semantic-zone support.}
$s_{\mathrm{zone}}$ measures whether the candidate pair appears in or near
query-relevant traffic regions, such as crosswalks, vehicle lanes, bike lanes,
turning areas, curb areas, or candidate conflict zones. If the query region is
ambiguous, the spatial uncertainty is stored in memory and passed to
verification.

\paragraph{Ground-plane proximity.}
$s_{\mathrm{prox}}$ measures whether the pair becomes close in the smoothed
ground-plane trajectory representation. Proximity is interpreted relative to the
participant classes and local traffic context. This term supports candidate
retrieval, but proximity alone does not establish a conflict.

\paragraph{Relative motion.}
$s_{\mathrm{motion}}$ measures whether the pair shows closing, converging,
crossing, or merging motion in trajectory space. Stationary co-presence,
parallel movement, or non-closing motion receives lower support.

\paragraph{Path or zone overlap.}
$s_{\mathrm{overlap}}$ measures whether the participants' trajectories,
extrapolated paths, or semantic zones overlap in a way that could create an
interaction.

\paragraph{Track reliability.}
$s_{\mathrm{track}}$ measures track continuity, common observation length, and
absence of severe identity switches or implausible jumps. Low reliability
reduces the candidate rank and is retained as uncertainty if the candidate is
later verified.

Together, these terms provide a traffic-specific retrieval signal. They are
intentionally permissive: a high coarse score means that an interval deserves
inspection, not that a conflict has been confirmed.

\subsubsection{Evidence Triage and Route Revision}
\label{app:diagnosis_replanning}

After candidate localization, the Traffic-Aware Controller inspects the
evidence buffer $\mathcal{M}^{evid}$ and decides whether to verify, expand,
tighten, or stop. This step makes the analysis adaptive: the route is revised
based on what the system has already found, rather than applying the same
processing to every interval. The corresponding evidence-triage prompt is
provided in Appendix~\ref{app:evidence_triage_prompt}.

\paragraph{Sparse evidence.}
If no plausible candidates are found, the controller may expand the temporal
window, relax an uncertain spatial filter, include additional participant-pair
types, or lower the coarse retrieval threshold within the query budget.

\paragraph{Excessive candidates.}
If too many candidates are found, the controller may tighten the spatial route,
increase track-quality requirements, or retain only the strongest evidence peaks
for verification.

\paragraph{Unreliable tracks.}
If candidates are dominated by identity switches, short tracks, or implausible
jumps, the controller may deprioritize those intervals or carry track
uncertainty into verification.

\paragraph{Ambiguous spatial grounding.}
If the query region is uncertain, the controller may preserve both a
grounded-zone route and a full-scene route. Spatial uncertainty is stored and
reported in the final uncertainty field.

\paragraph{Incomplete verification.}
If verification confirms only part of the requested evidence, the controller may
return to conflict-candidate localization with expanded windows, revised
filters, or additional participant-pair hypotheses.

\subsubsection{Fine Conflict Verification}
\label{app:Verify}

The fine conflict specialist is invoked only for candidate intervals selected
by the Traffic-Aware Controller. It reads the candidate participant pair,
grounded semantic region, tracks, trajectory cues, sampled visual evidence,
coarse score, and scene priors from memory. Whereas the coarse scanner ranks
plausibility, the specialist makes the final conflict decision.

Road users are detected and tracked in each frame. Tracks are cleaned and
smoothed, and each analysis window is represented by discrete ground-plane
trajectories. For road user $i$, the trajectory is
$\mathbf{p}^{\,i}_t \in \mathbb{R}^2$ with class label $c_i$ and
finite-difference velocity $\mathbf{v}^{\,i}_t$. All spatial quantities are
measured on the approximate ground plane.

\paragraph{Pair enumeration and minimum separation.}
For each unordered pair $(i,j)$ with sufficient common observation, we compute
the ground-plane separation series, minimum separation, and closest-approach
time:
\begin{equation}
\begin{aligned}
d_{ij}(t)
&=
\big\lVert
\mathbf{p}^{\,i}_t-\mathbf{p}^{\,j}_t
\big\rVert,\\
d^{\min}_{ij}
&=
\min_t d_{ij}(t),
\qquad
t^\ast
=
\arg\min_t d_{ij}(t).
\end{aligned}
\label{eq:min_separation}
\end{equation}
Here $d^{\min}_{ij}$ is the minimum separation between the two participants
over their common observation interval, measured on the approximate ground
plane. Pedestrian--pedestrian pairs are excluded because the target task is
road-user conflict analysis rather than crowd interaction analysis.

\paragraph{Closing-motion corroboration.}
TTC is evaluated at $t^\ast$ under a constant-velocity assumption. With
relative position
$\mathbf{r}=\mathbf{p}^{\,j}_{t^\ast}-\mathbf{p}^{\,i}_{t^\ast}$
and relative velocity
$\mathbf{v}=\mathbf{v}^{\,j}_{t^\ast}-\mathbf{v}^{\,i}_{t^\ast}$,
\begin{equation}
\mathrm{TTC}_{ij} =
\begin{cases}
\dfrac{-\,\mathbf{r}^{\top}\mathbf{v}}
{\lVert \mathbf{v} \rVert^{2}}, &
\mathbf{r}^{\top}\mathbf{v} < 0
\quad (\text{closing}),\\[6pt]
\text{undefined}, & \text{otherwise}.
\end{cases}
\label{eq:ttc}
\end{equation}
TTC is computed only for genuinely closing pairs and is used as corroborating
evidence, not as a stand-alone conflict decision. An event is marked
TTC-critical when
$\mathrm{TTC}_{ij}<\tau^{\mathrm{TTC}}=1.5$\,s.

\paragraph{Context-adaptive proximity threshold.}
A fixed proximity threshold is insufficient because the same ground-plane
separation can have different safety meaning depending on road-user class,
closing speed, approach geometry, local crowding, and visibility. We therefore
use a context-adaptive minimum-separation threshold:
\begin{equation}
\begin{aligned}
\tau^{d}_{ij}
=
\tau^{d}_{0}(c_i,c_j)
\cdot
\underbrace{f_{\mathrm{speed}}(v_{\mathrm{rel}})}
_{\text{closing speed}}
\cdot
\underbrace{f_{\mathrm{geom}}(\theta)}
_{\text{approach geometry}}
\cdot
\underbrace{f_{\mathrm{dens}}(\rho)}
_{\text{local density}}
\cdot
\underbrace{\phi_{\mathrm{vis}}}
_{\text{visibility}} .
\end{aligned}
\label{eq:adaptive_proximity}
\end{equation}
Here $\tau^{d}_{0}(c_i,c_j)$ is a base minimum-separation threshold for the
participant classes. The relative closing speed is
\begin{equation}
v_{\mathrm{rel}} =
\max\!\left(
0,
-\frac{\mathbf{r}^{\top}\mathbf{v}}
{\lVert\mathbf{r}\rVert}
\right),
\end{equation}
the approach angle is
\begin{equation}
\theta =
\arccos\!\left(
\frac{\mathbf{v}^{\,i\top}\mathbf{v}^{\,j}}
{\lVert\mathbf{v}^{\,i}\rVert
 \lVert\mathbf{v}^{\,j}\rVert}
\right),
\end{equation}
and $\rho$ is the number of nearby agents around the pair midpoint at
$t^\ast$. Each factor is bounded and monotone. The threshold widens for faster
closing interactions, adapts to approach geometry, accounts for crowded local
traffic, and can be adjusted under reduced visibility. Each factor can be
disabled for fixed-versus-adaptive comparisons and ablation studies.

\paragraph{Decision and severity.}
A candidate pair is confirmed only when the trajectory, motion, scene-region,
and response evidence jointly support a participant-level conflict. The primary
geometric condition compares the minimum ground-plane separation against the
context-adaptive threshold. Severity is graded by the fraction of the threshold
attained:
\begin{equation}
\mathrm{sev}_{ij}
=
\begin{cases}
\text{critical}, &
d^{\min}_{ij}
<
\gamma_{\mathrm{c}}\tau^{d}_{ij},\\

\text{high}, &
\gamma_{\mathrm{c}}\tau^{d}_{ij}
\le
d^{\min}_{ij}
<
\gamma_{\mathrm{h}}\tau^{d}_{ij},\\

\text{medium}, &
\gamma_{\mathrm{h}}\tau^{d}_{ij}
\le
d^{\min}_{ij}
<
\tau^{d}_{ij}.
\end{cases}
\label{eq:severity}
\end{equation}
where $\gamma_{\mathrm{c}} < \gamma_{\mathrm{h}} < 1$. Confirmed events are
written with participant identities, event time $t^\ast$, refined interval,
TTC flag, severity, response signatures, and conflict type. Rejected pairs are
not emitted as conflicts but remain countable through the audit trail.

\paragraph{Specialist decision record.}
The specialist first assembles the candidate pair's trajectories over the
flagged interval and applies track-quality gates. Candidates are rejected when
the supporting tracks contain severe identity switches, implausible jumps,
insufficient common observation, or unreliable localization. Surviving
candidates are evaluated using minimum ground-plane separation, the
context-adaptive threshold of Eq.~\eqref{eq:adaptive_proximity}, closing or
converging motion, semantic-zone consistency, trajectory overlap, TTC evidence
when available, and visible road-user response.

Each decision includes an uncertainty assessment stored as $u_i$. This
uncertainty summarizes the separation margin relative to the contextual
threshold, agreement between proximity and TTC evidence, track reliability,
semantic-zone confidence, and visual support. Confirmed candidates are written
to reasoning-result memory with their refined interval, participant identities,
semantic region, conflict type, primary responses, trajectory and motion
evidence, uncertainty, and clip information. Rejected candidates are written
with explicit rejection reasons, such as ordinary co-presence, non-closing
motion, insufficient track quality, separation above the contextual threshold,
ambiguous spatial support, or lack of visible response.

\paragraph{Severity-weighted risk field.}
Each confirmed event contributes a severity-weighted point
$\big(\mathbf{x}_e,s_e\big)$, where
$\mathbf{x}_e=\tfrac{1}{2}
(\mathbf{p}^{\,i}_{t^\ast}+\mathbf{p}^{\,j}_{t^\ast})$
is the event location and $s_e$ is the severity weight. The event set is
accumulated on a regular grid and smoothed with an isotropic Gaussian kernel:
\begin{equation}
\begin{aligned}
D(\mathbf{x})
&=
\left(
G_\sigma *
\textstyle\sum_e
s_e\delta_{\mathbf{x}_e}
\right)(\mathbf{x}),\\
\tilde{D}(\mathbf{x})
&=
\frac{D(\mathbf{x})}
{\max_{\mathbf{x}}D(\mathbf{x})}.
\end{aligned}
\label{eq:density}
\end{equation}
This field converts sparse verified events into interpretable hotspot zones
while weighting severe interactions more strongly than frequent weak ones.

\subsubsection{Conflict Typology and Kinematic Response Signatures}
\label{app:conflict_typology}

The fine conflict specialist assigns each verified event an automatic conflict
type and per-participant response signatures. The automatic vocabulary follows
the same broad conflict categories used by the human annotation protocol, while
the labels here are computed from trajectory and response evidence and should
therefore be interpreted as automatic kinematic classifications.

\paragraph{Conflict detection criterion.}
Following the traffic-conflict tradition, we treat a conflict as an observable
interaction in which two road users approach each other in space and time such
that a collision could occur if their movements remained unchanged
~\cite{amundsen1977traffic,hyden1987development}. Conventional surrogate-safety
analysis includes temporal measures such as Post-Encroachment Time (PET) and
Time-to-Collision (TTC)
~\cite{allen1978analysis,gettman2003surrogate,gettman2008surrogate}.
In our implementation, the primary geometric quantity is the minimum
ground-plane separation $d^{\min}_{ij}$ in
Eq.~\eqref{eq:min_separation}, evaluated against the context-adaptive proximity
threshold in Eq.~\eqref{eq:adaptive_proximity}. TTC is used as a corroborating
temporal measure and is computed using Eq.~\eqref{eq:ttc}, following Hayward's
time-to-collision formulation~\cite{hayward1972near}. TTC is reported only when
the pair is closing, and an event is marked TTC-critical when
$\mathrm{TTC}<1.5$\,s, consistent with commonly used urban surrogate-safety
thresholds~\cite{gettman2003surrogate,gettman2008surrogate}. TTC alone does
not confirm a conflict; it must agree with trajectory, semantic-region, and
road-user response evidence.

Pedestrian--pedestrian pairs are excluded because the focus is on road-user
conflicts rather than crowd interactions. Motorcycles are excluded when they
are not covered by the upstream detector class set. All distances, speeds, and
response signatures used by the specialist are computed from projected
ground-plane trajectories, supporting metric comparison within each fixed
camera view.

\paragraph{Type assignment.}
Each verified event receives one automatic label from the conflict ontology.
Rules are applied in priority order; the first matching rule determines the
reported conflict type.

\begin{enumerate}
    \item \textbf{Yielding.}
    At least one participant exhibits a slowing response approaching closest
    approach, including braking, stopping, or already stopped. Braking is
    defined as a speed drop exceeding $30\%$. The record retains which
    participant yielded. Behavioral response takes precedence over geometry,
    consistent with conflict-observation methods that emphasize evasive or
    corrective road-user actions
    ~\cite{hyden1987development,parker1989traffic}.

    \item \textbf{Turning-path.}
    At least one participant executes a turn, defined as total heading change
    across the interaction window exceeding $\pm25^{\circ}$.

    \item \textbf{Crossing-path.}
    A cross-type pair, such as vehicle--pedestrian, vehicle--bicycle, or
    bicycle--pedestrian, proceeds through intersecting or competing paths. This
    follows the intuition of angle-based surrogate-safety typologies, where
    conflicts are distinguished by the relative geometry of the interacting
    paths
    ~\cite{gettman2003surrogate,gettman2008surrogate,ssam2008manual}.

    \item \textbf{Merging--following.}
    A same-type pair proceeds through a shared, converging, or following path.
    This covers interactions analogous to rear-end, following, merging, and
    lane-change conflicts.

    \item \textbf{Close-proximity.}
    The pair satisfies the adaptive proximity condition, but the available
    trajectory and response evidence does not support a more specific kinematic
    mechanism. This includes stationary or partially observed participants.
\end{enumerate}

Our maneuver rules use each participant's own heading change and response
signature, whereas SSAM-style typologies often classify conflicts using the
angle between interacting paths or vehicles
~\cite{gettman2008surrogate,ssam2008manual}. The two conventions agree on
prototypical cases but are not identical. \emph{Gap acceptance} is not emitted
by the automatic classifier because it is not reliably inferable from the
available sampled trajectories.

\paragraph{Response signatures.}
For each participant, we record the kinematic counterpart of the observable
response field. Signatures are computed from the trajectory in a short window
around closest approach.

\begin{table*}[t]
\centering
\caption{Per-participant response signatures and decision rules.}
\label{tab:response_rules}
\small
\begin{tabular}{@{}lll@{}}
\toprule
\textbf{Signature} & \textbf{Rule} & \textbf{Threshold} \\
\midrule
braking
& speed drop approaching closest approach
& $>30\%$ \\

stopping / stopped
& speed below floor, either now or already
& $v_{\mathrm{stop}}$ \\

accelerating
& speed rise approaching closest approach
& $>40\%$ \\

directional adjustment
& heading change around the encounter
& $>20^{\circ}$ \\

evasive motion
& either participant adjusts direction
& --- \\

yielding
& one participant slows while the other maintains motion
& --- \\
\bottomrule
\end{tabular}
\end{table*}

\paragraph{Parameter choices and validation status.}
All numerical parameters above are defaults chosen by inspection,
sanity-checked against trajectory-level reconstructions of the recordings, and
applied uniformly across all sites and datasets. They were not tuned against
ground truth. The human-annotated conflict records use the corresponding
conflict and response vocabulary, enabling direct field-by-field agreement
measurement between automatic outputs and human annotations.
\subsubsection{Adaptive Stopping and Audit Trail}
\label{app:adaptive_stopping_details}

The Traffic-Aware Controller stops when the workspace indicates that the query
has been answered or that further search is unlikely to change the result.
Stopping can occur when verified records satisfy the query requirements with
acceptable uncertainty, when all plausible candidates have been verified or
rejected, when additional route revisions produce no new high-quality evidence,
or when the query-specific budget is exhausted. The memory-update prompt used
to decide whether to stop or continue is provided in
Appendix~\ref{app:memory_update_prompt}.

For the conflict queries used in the main evaluation, the system
outputs verified participant-level conflict records. If the system cannot verify
a query-matching conflict within the allotted budget, it reports the searched
windows, relevant regions, candidate counts, rejection reasons, and remaining
uncertainty. In all cases, the output preserves an audit trail from the original
query to the selected route, flagged intervals, verification decisions, and
final records.

\subsection{Prompt Details}
\label{app:prompt_details}

Prompted reasoning in \textsc{AgenticConflictLens} is used only for
language-dependent and controller-level decisions. It is not used to compute
geometric conflict evidence. Proximity, relative motion, path overlap, TTC,
track-quality checks, response signatures, and uncertainty quantities are
computed directly from detections, tracks, smoothed ground-plane trajectories,
and scene-prior memory. Prompts are used for query parsing, scene-aware route
planning, evidence triage, record and memory update, and final record
generation.

\subsubsection{Query Parser Prompt}
\label{app:query_parser_prompt}

\begin{promptbox}{Query Parser Prompt}
Given a natural-language traffic-conflict query, extract the structured task
state.

Identify the temporal scope, spatial scope, road-user classes,
participant-pair types, interaction type, and evidence requirements. If a field
is not specified, mark it as unspecified rather than guessing.

Return the time range, scene region, participant classes, candidate pair types,
conflict type, required evidence, query confidence, and unresolved ambiguities.
\end{promptbox}

\subsubsection{Scene-Aware Route-Planning Prompt}
\label{app:route_planning_prompt}

\begin{promptbox}{Scene-Aware Route-Planning Prompt}
Given the parsed task-state memory and scene-prior memory, choose an analysis
route for the query.

Select the temporal windows, semantic traffic regions, road-user classes,
participant-pair types, and evidence requirements that should be used by
conflict-candidate localization.

If the query is focused, create a narrow route over the specified time or
region. If the query is open-ended, create a broader route using relevant scene
priors and participant-pair hypotheses.

Return the selected route, the reason for the route, and the evidence-buffer
fields that should be updated.
\end{promptbox}

\subsubsection{Evidence-Triage Prompt}
\label{app:evidence_triage_prompt}

\begin{promptbox}{Evidence-Triage Prompt}
Given the candidate intervals, coarse scores, semantic-zone support,
participant-pair evidence, track-quality indicators, trajectory cues, and
uncertainty flags, choose the next controller action.

Select one action: verify candidates, expand the temporal window, relax an
uncertain spatial filter, include additional participant-pair types, tighten
the spatial route, increase track-quality requirements, keep only the strongest
evidence peaks, or stop with a negative or uncertain result.

Return the chosen action, the reason, and the memory fields that should be
updated.
\end{promptbox}

\subsubsection{Record and Memory-Update Prompt}
\label{app:memory_update_prompt}

\begin{promptbox}{Record and Memory-Update Prompt}
Given the verification result for each flagged interval, update the
reasoning-result memory.

For confirmed candidates, store the refined interval, participant pair,
semantic region, conflict type, road-user responses, trajectory evidence,
uncertainty, and supporting clip identifier.

For rejected candidates, store the rejection reason and any uncertainty that
affected the decision.

Decide whether the query has been answered or whether the controller should
return to scene-aware route planning or conflict-candidate localization.

Return the updated memory fields and the next controller decision.
\end{promptbox}

\subsubsection{Record-Generator Prompt}
\label{app:record_generator_prompt}

\begin{promptbox}{Record-Generator Prompt}
Generate final conflict records from the verified results stored in memory.

For each verified event, report the refined time interval, participant pair,
semantic traffic region, conflict type, observable road-user responses,
trajectory evidence, uncertainty, and supporting clip identifier.

Also summarize rejected candidate intervals with their rejection reasons.

Do not invent events, responses, conflict types, uncertainty values, or clip
identifiers that are not stored in memory.
\end{promptbox}

\providecommand{\method}{}
\renewcommand{\method}{\textsc{AgenticConflictLens}}

\section{Extended Experimental Details}
\label{app:experiments}

This appendix provides the extended experimental protocol and additional
results for \method. 
The organization mirrors Sec.~4: datasets, evaluation
protocol and implementation, main query-driven results,
ablation studies, and additional spatial-risk analyses
on the real-world traffic recordings.
The
protocol is designed to evaluate the full workflow defined in
Sec.3: the Traffic-Aware Controller, structured traffic memory,
reusable fixed-camera scene priors, homography-projected trajectories,
specialized traffic-analysis tools, coarse conflict scanning, fine conflict
verification, and auditable participant-level record generation.

\subsection{Conflict Annotations and Query Construction}
\label{app:annotation_and_queries}

\subsubsection{Conflict Record Definition}
\label{app:conflict_record_definition}

A road-user conflict is a temporally localized encounter in which two or more
road users exhibit converging, competing, or mutually dependent movement, with
visible evidence such as yielding, braking, stopping, directional adjustment,
gap acceptance, or evasive motion. The annotation is based only on observable
motion and response. Proximity alone is not sufficient: a valid conflict must
include trajectory, motion, scene-region, and response evidence consistent with
the verification criteria used by the fine conflict specialist.

\paragraph{Temporal boundaries.}
The start boundary is the earliest time at which conflict-related movement or
response becomes visible. Depending on the case, this can correspond to the
beginning of path convergence, visible deceleration or stopping, directional
adjustment, entry into another participant's projected path, initiation of a
turning or merging response, or another observable motion change associated with
the conflict. The end boundary is the time at which the participants' paths
separate, the visible response is completed, or the conflict-related movement is
no longer observable. A continuous encounter involving the same participants is
annotated once, even when it spans several automatically proposed candidate
windows.

\paragraph{Conflict record fields.}
Each verified conflict record contains the video identifier, conflict identifier,
start and end timestamps, participant pair, conflict type, primary observable
response, secondary observable response when present, relevant scene region, and
a concise evidence statement. A representative record is shown below.

\begin{verbatim}
{
  "video_id": "BELLEVUE_03_004",
  "conflict_id": "CONFLICT_0092",
  "start_time_sec": 763.0,
  "end_time_sec": 772.0,
  "participant_pair": ["auto", "ped"],
  "conflict_type": "turning_path",
  "response": "auto_slows",
  "scene_region": "crosswalk",
  "evidence":
    "The turning auto slows after the ped
     enters its projected path."
}
\end{verbatim}

\paragraph{Participant families.}
We group participant pairs into four broad families: auto--pedestrian,
auto--cyclist, auto--auto, and conflicts involving a bus or truck.

\paragraph{Conflict ontology.}
We use a compact ontology based on visually observable motion and response:

\begin{enumerate}
    \item \textbf{Yielding conflict}: one participant visibly slows or stops
    in response to another participant's movement or projected path;

    \item \textbf{Turning-path conflict}: a turning participant intersects,
    approaches, or responds to another participant's path;

    \item \textbf{Crossing-path conflict}: participants approach or move
    through intersecting trajectories and at least one exhibits a visible
    response or path adjustment;

    \item \textbf{Merging/following conflict}: a lane-entry, following,
    gap-acceptance, or related longitudinal encounter produces visible
    movement dependence between participants; and

    \item \textbf{Close-proximity/evasive conflict}: a participant exhibits
    braking, swerving, stopping, or another visible evasive response during a
    close or converging encounter.
\end{enumerate}

These categories describe visually observable conflict patterns only and do not
assign fault, intent, or right-of-way.

\begin{table*}[t]
\centering
\small
\caption{Conflict and query statistics for the
96-hour real-world evaluation, comprising the
Bellevue and NYC DOT datasets.
}
\label{tab:annotation_statistics_app}
\begin{tabular}{lccc}
\toprule
\textbf{Evaluation source}
& \textbf{Hours}
& \textbf{Conflicts}
& \textbf{Queries} \\
\midrule
Bellevue public evaluation
& 60 & 119 & 457 \\
NYC DOT annotated evaluation
& 36 & 1008 & 4125 \\
\midrule
Total annotated evaluation
& 96 & 1127 & 4582 \\
\bottomrule
\end{tabular}
\end{table*}

\subsubsection{Annotation Construction and Quality Control}
\label{app:annotation_quality}

Human annotators review the traffic recordings and identify road-user conflicts
according to the operational definition and conflict ontology. Detector and
tracker outputs may be used as visual aids for locating road users and
reviewing temporal context, but they do not determine whether a conflict is
present and are not used as ground-truth labels.

For each verified conflict, annotators record the start and end timestamps,
participant pair, conflict type, visible response, relevant scene region,
and a concise evidence statement. Temporal boundaries span from the earliest
observable conflict-related behavior to the point where the participants' paths
separate or the visible response is completed. Continuous conflicts involving
the same participants are annotated once, and duplicate records are removed.

A second annotator independently verifies the Bellevue and annotated NYC DOT
conflict annotations using the same guidelines. Verification covers conflict
validity, participant pair, conflict type, temporal boundaries, visible
response, and interval completeness. Disagreements are resolved by a third
reviewer, and the adjudicated annotation is used as the final reference. We
measure pre-adjudication consistency using participant-pair agreement,
conflict-type agreement, Cohen's $\kappa$ for conflict-type labels,
temporal IoU, and absolute start- and end-boundary differences.

Queries are also manually reviewed for answerability, temporal scope,
participant constraints, consistency with the reference conflict, and ambiguity.
Accepted queries are used directly, corrected queries are revised before use,
and ambiguous queries are removed from the evaluation.

\begin{table}[t]
\centering
\small
\caption{Annotation and query quality-control results. Agreement and temporal
boundary measures are computed before adjudication; query results summarize the
manual query-review process.}
\label{tab:annotation_quality_app}
\begin{tabular}{lc}
\toprule
\textbf{Measure} & \textbf{Result} \\
\midrule
Participant-pair agreement & 93.4\% \\
Conflict-type agreement & 88.7\% \\
Conflict-type Cohen's $\kappa$ & 0.84 \\
Mean temporal IoU & 0.79 \\
Median start-boundary difference & 1.3 s \\
Median end-boundary difference & 1.6 s \\
Query acceptance without revision & 83.6\% \\
Queries corrected during review & 13.1\% \\
Queries rejected as ambiguous & 3.3\% \\
\bottomrule
\end{tabular}
\end{table}

\subsubsection{Query Construction}
\label{app:query_details}

Queries are constructed only after the corresponding conflict annotations have
been verified. Each verified conflict is first represented as a canonical
structured record compatible with the method output: refined interval,
participant pair, semantic scene region, conflict type, observable response,
trajectory evidence, uncertainty note when relevant, and supporting evidence
statement. Controlled templates are then used to generate candidate
natural-language queries and reference answers, followed by manual review.

The query-generation stage does not modify the underlying conflict annotation.
Reference timestamps, participant labels, conflict types, scene regions,
and visible responses are taken directly from the verified annotation rather
than inferred from system predictions.

Each query specification stores the video identifier, natural-language query,
temporal search scope, participant constraints, optional scene-region
constraint, optional maneuver or conflict constraint, aggregation requirement
such as \emph{first}, \emph{count}, or \emph{all}, expected answer format, and
associated conflict identifiers.

\paragraph{Query families.}
We organize the 4,582 real-world evaluation queries into five families that
test different aspects of query-conditioned long-video conflict search.

\paragraph{Temporally scoped retrieval.}
These queries provide a temporal range together with participant or spatial
constraints and require retrieval of the matching conflict within that interval.
\begin{quote}
\emph{Between 10:00 and 18:20, identify the auto--pedestrian conflict near the
crosswalk.}
\end{quote}

\paragraph{Full-video actor-conditioned search.}
These queries specify participant categories but do not reveal the target
conflict time, requiring search over the complete video.
\begin{quote}
\emph{Identify the auto--cyclist conflict in the complete video.}
\end{quote}

\paragraph{Conflict confirmation.}
These queries ask whether a specified conflict pattern occurs in the video or
within a temporal scope. Each reference answer is tied to at least one verified
conflict record.
\begin{quote}
\emph{Does a bus--cyclist conflict occur in the video?}
\end{quote}

\paragraph{Localization and reasoning.}
These queries require either temporal grounding of a conflict or interpretation
of directly observable behavior within a localized conflict.
\begin{quote}
\emph{When does the first conflict between a turning auto and a pedestrian
occur?}
\end{quote}
\begin{quote}
\emph{How does the auto respond after the pedestrian enters its path?}
\end{quote}

\paragraph{Open-ended search.}
These queries omit both the conflict time and participant pair, requiring broad
video search before local conflict interpretation.
\begin{quote}
\emph{Identify the first road-user conflict in the video and report its
timestamp, participants, conflict type, and visible response.}
\end{quote}

\paragraph{Query distribution.}
Table~\ref{tab:query_distribution_app} reports the query-family composition.
The counts sum to 4,582 real-world evaluation queries. Queries in the
full-video actor-conditioned and open-ended families do not reveal the target
timestamp; additional confirmation and localization/reasoning queries also omit
the timestamp, so at least 50\% of the evaluation queries require long-video
search rather than only local clip interpretation.

\begin{table}[t]
\centering
\small
\caption{Query-family composition for the 4,582 real-world evaluation
queries.}
\label{tab:query_distribution_app}
\begin{tabular}{lcc}
\toprule
\textbf{Family} & \textbf{Number} & \textbf{Share} \\
\midrule
Temporally scoped retrieval & 1040 & 22.7\% \\
Full-video actor-conditioned search & 1251 & 27.3\% \\
Conflict confirmation & 898 & 19.6\% \\
Localization and reasoning & 935 & 20.4\% \\
Open-ended search & 458 & 10.0\% \\
\bottomrule
\end{tabular}
\end{table}

\paragraph{Query difficulty.}
For analysis, we group queries into three levels according to how much
localization information is provided:

\begin{itemize}
    \item \textbf{Scoped}: the query provides both a temporal range and
    participant constraints;

    \item \textbf{Actor-conditioned}: the query provides participant
    constraints but does not reveal the conflict time; and

    \item \textbf{Open-ended}: the query provides neither the conflict time nor
    the participant pair.
\end{itemize}

These difficulty groups are used only for evaluation and do not change the
inference procedure, prompts, thresholds, controller budget, or stopping
criteria.

\paragraph{Query verification.}
Each evaluation query is manually reviewed for grammatical clarity, consistency
with the source annotation, correctness of its temporal scope, answerability
from visible evidence, and agreement between the requested constraints and
reference answer. Queries requiring unsupported judgments about intent, fault,
causality, or right-of-way are revised or excluded.

\subsection{Detailed Baseline Implementation and Metrics}
\label{app:baseline_metric_details}

This appendix provides the detailed baseline implementations, shared
post-processing rules, and metric definitions used in
Sec.4.2. Each adapted baseline is evaluated under
the same conflict-localization interface: given a long traffic-camera video and
a natural-language query, the method returns candidate conflict intervals and
participant pairs.

\begin{table*}[t]
\centering
\small
\caption{Adapted baselines for long-video conflict localization. Each method is
evaluated through the same practical interface: returning candidate conflict
intervals and participant pairs for comparison with annotated conflict records.}
\label{tab:baseline_comparison_app}
\begin{tabular}{p{0.26\linewidth} p{0.36\linewidth} p{0.28\linewidth}}
\toprule
\textbf{Method} & \textbf{Returned output for comparison} & \textbf{How it is compared} \\
\midrule

Top retrieved conflict clips~\cite{wang2025videotree} &
Top retrieved conflict clips from hierarchical coarse-to-fine search &
tIoU, pair accuracy, efficiency \\

Uniform clip-level conflict search~\cite{zhang2025seeunsafe} &
Clips uniformly searched and judged unsafe/conflict by VLM &
tIoU, pair accuracy, efficiency \\

Memory-selected conflict clips~\cite{yin2026videoarm} &
Conflict clips selected from long-video memory entries &
tIoU, pair accuracy, efficiency \\

Tool-selected conflict clips~\cite{lin2026videoseek} &
Conflict clips selected through tool-guided temporal seeking &
tIoU, pair accuracy, efficiency \\

High-risk trajectory intervals~\cite{ssam2008manual} &
Trajectory intervals ranked by surrogate-risk evidence &
tIoU, pair accuracy, efficiency \\

\method{} &
Verified refined conflict intervals and participant-level records &
tIoU, pair accuracy, efficiency \\
\bottomrule
\end{tabular}
\end{table*}

\subsubsection{Shared Evaluation Interface}
\label{app:shared_eval_interface}

All methods receive the same video, natural-language conflict query, timestamp
convention, participant vocabulary, candidate-output format, and query-time
budget when applicable. Each method returns a ranked list of predicted temporal
intervals and participant pairs.

Unless otherwise stated, videos are decoded at 4 fps for analysis. For
trajectory-based components, road users are detected with YOLOv8x
(\texttt{imgsz}=1536, confidence threshold 0.20) and tracked with ByteTrack.
Tracks are smoothed and, when camera calibration is available, projected to an
approximate ground plane using a fixed-view homography. All thresholds and
hyperparameters are fixed before evaluation and applied unchanged across
Bellevue and NYC DOT.

All VLM-based baselines use the same frozen multimodal backbone as \method{},
Qwen2.5-VL-7B-Instruct. The model is not fine-tuned. We use fixed prompts for
clip relevance, unsafe-interaction judgment, temporal evidence selection, and
participant-pair extraction. The same prompts are used for Bellevue and NYC DOT
and are not tuned on evaluation annotations. Each VLM call returns a structured
JSON object containing a binary decision, a confidence score in $[0,1]$, the
predicted participant pair, a short evidence statement, and the clip timestamp.
Clips with confidence below 0.5 are rejected unless they are used only for
ranking inside a hierarchical search baseline.

\subsubsection{Baseline Implementations}
\label{app:baseline_implementation_details}

\paragraph{VideoTree-inspired hierarchical retrieval.}
This baseline evaluates whether a generic hierarchical long-video retrieval
strategy can find the conflict interval. Each video is first divided into coarse
150-second chunks. For each chunk, we uniformly sample frames and ask the VLM to
score whether the chunk may contain evidence relevant to the query. The
top-ranked coarse chunks are then expanded into 45-second candidate windows with
30-second stride. Each candidate window is rescored by the VLM using the same
conflict-relevance prompt. The highest-scoring fine windows are returned as
\emph{top retrieved conflict clips}. Participant pairs are extracted from the
selected clips using the shared participant-pair extraction prompt. This
baseline uses hierarchical temporal expansion only; it does not use
traffic-scene memory, trajectory-risk features, or fine conflict verification.

\paragraph{SeeUnsafe-inspired accident-aware VLM search.}
This baseline evaluates whether direct VLM-based unsafe-event recognition can
localize the requested conflict. Each video is divided into fixed 10-second
clips. For each clip, the frozen VLM receives the clip and query and is asked
whether the clip contains the requested unsafe interaction, near miss, yielding,
braking, evasive motion, or close conflict, rather than ordinary co-presence.
Clips judged positive with confidence at least 0.5 are returned as
\emph{clips judged unsafe/conflict by VLM}. The participant pair is extracted
from each positive clip using the shared extraction prompt. This baseline does
not perform hierarchical search, memory construction, trajectory verification,
or route revision; it is a direct clip-level safety-VLM baseline.

\paragraph{VideoARM-inspired hierarchical-memory retrieval.}
This baseline evaluates whether long-video memory can support conflict
search. Each video is divided into 45-second memory segments with 30-second
stride. For each segment, the VLM generates a short memory entry describing
visible road users, scene regions, maneuvers, and possible interaction cues.
Given the query, the agent ranks the memory entries by relevance and visually
inspects the top-ranked entries. The inspected segments are rescored by the VLM
for conflict evidence, and the selected segments are returned as
\emph{memory-selected conflict clips}. Participant pairs are extracted from the
selected clips using the shared extraction prompt. This baseline uses 
temporal memory and VLM relevance scoring, but it does not use the structured
traffic memory, trajectory-risk gates, or fine conflict specialist used by
\method{}.

\paragraph{VideoSeek-inspired tool-guided seeking.}
This baseline evaluates whether a tool-guided seeking process can
efficiently jump to the relevant conflict interval. The agent is given three
generic actions: \texttt{overview}, \texttt{skim}, and \texttt{focus}.
\texttt{overview} scores 150-second coarse chunks; \texttt{skim} inspects
45-second windows inside promising chunks; and \texttt{focus} selects final
10-second clips around the strongest evidence. At each step, the VLM decides
whether to skip, inspect, revisit, or zoom into a temporal region based only on
the query and visual evidence. The final focused clips are returned as
\emph{tool-selected conflict clips}, and participant pairs are extracted with
the shared extraction prompt.

\paragraph{SSAM-inspired surrogate-risk search.}
This baseline evaluates how far a trajectory-only surrogate-safety method can
go under the same query-conditioned setting. We use the same detections,
ByteTrack tracks, and homography-projected trajectories as \method{}. The query
is converted into filters over participant class, temporal range, and scene
region when specified. For each candidate participant pair, the baseline
computes proximity, path overlap, closing motion, TTC/PET-style cues, and track
reliability over 1-second bins. Candidate intervals are scored as
\begin{equation}
\begin{aligned}
s_{\mathrm{SSAM}} =
&\,0.30s_{\mathrm{prox}} +
0.25s_{\mathrm{ttc/pet}} +
0.20s_{\mathrm{overlap}} \\
&+ 0.15s_{\mathrm{motion}} +
0.10s_{\mathrm{track}} .
\end{aligned}
\end{equation}
Bins with $s_{\mathrm{SSAM}}\geq0.5$ are retained, adjacent positive bins
separated by less than 2 seconds are merged, and the resulting intervals are
returned as \emph{high-risk trajectory intervals}. The participant pair is
obtained directly from the corresponding tracks. This baseline does not use VLM
visual reasoning for the final risk score.

\paragraph{\method{}.}
Our method uses the same shared detections, tracks, VLM backbone, and query
budget, but differs in how evidence is selected and verified. The
Traffic-Aware Controller parses the query, uses fixed-camera scene context and
structured traffic memory, retrieves candidate intervals with the traffic-aware
coarse scanner, and sends only selected intervals to the fine conflict
specialist. The fine specialist verifies each candidate using trajectory,
motion, semantic-region, visual-response, track-quality, and uncertainty
evidence. Only verified candidates are emitted as \emph{verified refined
conflict intervals} and written as participant-level conflict records.

\paragraph{Shared post-processing.}
For every method, predicted intervals are clipped to the query temporal scope,
overlapping intervals are merged, and intervals shorter than 1 second are
expanded to 1 second around their center for scoring. If a method returns
multiple intervals for the same physical event, only the highest-overlap match
is counted as a true positive; remaining intervals are counted as false
positives. Runtime, decoded frames, VLM calls, visual tokens, and inspection
percentage are logged for every query.
\paragraph{Computing environment.}
All experiments were conducted on Linux using three NVIDIA RTX A6000 GPUs
with 48\,GB of memory each, PyTorch 2.2.1, and CUDA 12.1.
\subsubsection{Metric Definitions}
\label{app:metric_details}

All methods are evaluated on the shared output used in
Table 1: predicted conflict intervals and participant
pairs. Let the reference set for query $q$ be
$\mathcal{G}_q=\{(I_j,p_j)\}_{j=1}^{N_q}$, where
$I_j=[t^s_j,t^e_j]$ is the annotated conflict interval and $p_j$ is the
annotated participant pair. Let the method predictions be
$\hat{\mathcal{G}}_q=\{(\hat{I}_k,\hat{p}_k)\}_{k=1}^{M_q}$.

\paragraph{Temporal localization.}
Temporal overlap between a predicted interval
$\hat{I}_k=[\hat{t}^s_k,\hat{t}^e_k]$ and a reference interval
$I_j=[t^s_j,t^e_j]$ is measured by temporal intersection-over-union:
\begin{equation}
\mathrm{tIoU}(\hat{I}_k,I_j)=
\frac{
\max\!\left(0,\min(\hat{t}^e_k,t^e_j)-\max(\hat{t}^s_k,t^s_j)\right)
}{
\max(\hat{t}^e_k,t^e_j)-\min(\hat{t}^s_k,t^s_j)
}.
\end{equation}
Predicted and reference intervals are matched one-to-one using maximum-overlap
bipartite matching. A reference conflict is considered retrieved when it is
matched by a prediction with $\mathrm{tIoU}\geq0.5$.

\paragraph{Conflict-existence F1.}
\textbf{Exist. F1} measures event-level conflict retrieval. After one-to-one
matching, true positives are temporally matched predictions, false positives are
unmatched predictions, and false negatives are unmatched reference conflicts:
\begin{equation}
\begin{aligned}
P &= \frac{TP}{TP+FP}, \qquad
R = \frac{TP}{TP+FN}, \\
\mathrm{Exist.\ F1} &=
\frac{2PR}{P+R}.
\end{aligned}
\end{equation}

Mean tIoU is computed over the matched true-positive predictions:
\begin{equation}
\overline{\mathrm{tIoU}} =
\frac{1}{TP}
\sum_{(\hat{I}_k,I_j)\in \mathcal{M}}
\mathrm{tIoU}(\hat{I}_k,I_j),
\end{equation}
where $\mathcal{M}$ is the set of matched true-positive interval pairs.

\paragraph{Participant-pair accuracy.}
Pair accuracy evaluates whether the method identifies the correct road-user
pair after localizing the conflict interval. It is computed over matched
true-positive intervals:
\begin{equation}
\mathrm{PairAcc} =
\frac{1}{|\mathcal{M}|}
\sum_{(\hat{I}_k,I_j)\in \mathcal{M}}
\mathbf{1}\!\left[\hat{p}_k=p_j\right].
\end{equation}
This separates temporal retrieval from participant-level correctness: a method
may retrieve the right time window while still identifying the wrong interacting
road users.

\paragraph{Structured-record accuracy.}
For \method{} and its ablated variants, we additionally evaluate the complete
participant-level conflict record. A verified record is counted as correct only
when it matches the required fields for the query: interval, participant pair,
semantic scene region, conflict type, and primary road-user response. Complete
structured-record accuracy is defined as
\begin{equation}
c_q^{\mathrm{record}} =
c_q^{\mathrm{time}}
c_q^{\mathrm{pair}}
c_q^{\mathrm{region}}
c_q^{\mathrm{type}}
c_q^{\mathrm{response}},
\end{equation}
where each term is binary and is omitted only when the query does not require
that field. This metric is used for \method{} and ablations, not for ranking
adapted baselines that do not natively emit the full record schema.

\paragraph{Efficiency.}
Efficiency measures how much video and multimodal computation each method uses
to answer a query. The inspection percentage is
\begin{equation}
\mathrm{Inspect\%}_q =
100 \times
\frac{
\left|\bigcup_{\hat{I}_k \in \hat{\mathcal{F}}_q}\hat{I}_k\right|
}{
t^q_{\mathrm{end}}-t^q_{\mathrm{start}}
},
\end{equation}
where $\hat{\mathcal{F}}_q$ is the set of intervals inspected by the final
candidate-selection or verification stage. For \method{}, these are the
intervals sent to the fine conflict specialist. For retrieval baselines, these
are the clips returned or inspected as selected candidate evidence.

We also report decoded frames, multimodal calls, visual tokens, and runtime.
Frames count the number of decoded video frames used during query processing.
Calls count VLM invocations. Tokens count visual tokens consumed by VLM-based
methods. Runtime measures query-time inference cost under the same hardware and
budget setting. All metrics are averaged over evaluation queries and reported
separately for Bellevue and NYC DOT.

\paragraph{Coarse-to-fine verification.}
To evaluate the traffic-aware coarse scanner and fine specialist separately, we
report the number of coarse flagged intervals, the number verified as true
conflicts, and the rejection rate:
\begin{equation}
\mathrm{RejectRate} =
100 \times
\frac{N_{\mathrm{flagged}} - N_{\mathrm{verified}}}
{N_{\mathrm{flagged}}}.
\end{equation}
A high rejection rate is expected when the scanner favors recall; it is useful
when true conflicts are retained while ordinary co-presence, weak interactions,
ambiguous spatial support, and noisy tracks are filtered before final record
generation.

\subsection{Qualitative Case Studies and Failure Analysis}
\label{app:qualitative_case_studies}

To complement the aggregate results, we show two end-to-end traces illustrating
a successful retrieval and a representative failure mode of \method{}.

\begin{figure*}[t]
    \centering
    \includegraphics[width=1\linewidth]{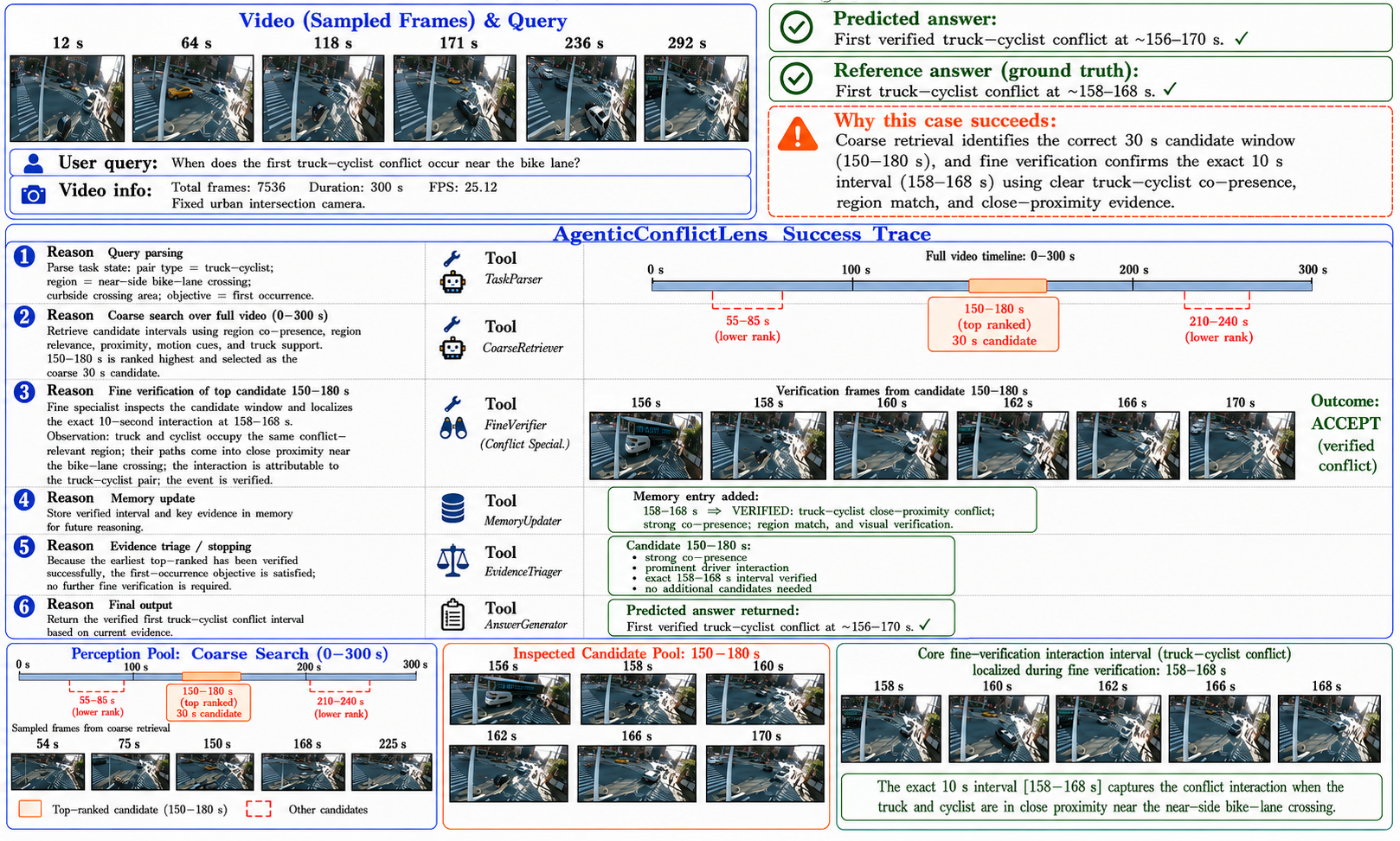}
    \caption{Success case: coarse retrieval selects 150--180~s and fine
    verification localizes the verified truck--cyclist conflict to
    approximately 158--168~s near the bike lane.}
    \label{fig:Success_Case}
\end{figure*}

\paragraph{Success case.}
Figure~\ref{fig:Success_Case} shows a truck--cyclist query for which coarse
retrieval selects the 150--180~s candidate from the 300-s video. Fine
verification then localizes the interaction to approximately 158--168~s near
the bike lane and confirms the conflict. The verified interval and supporting
evidence are stored in memory, and the returned answer closely matches the
reference.

\begin{figure*}[t]
    \centering
    \includegraphics[width=0.9\linewidth]{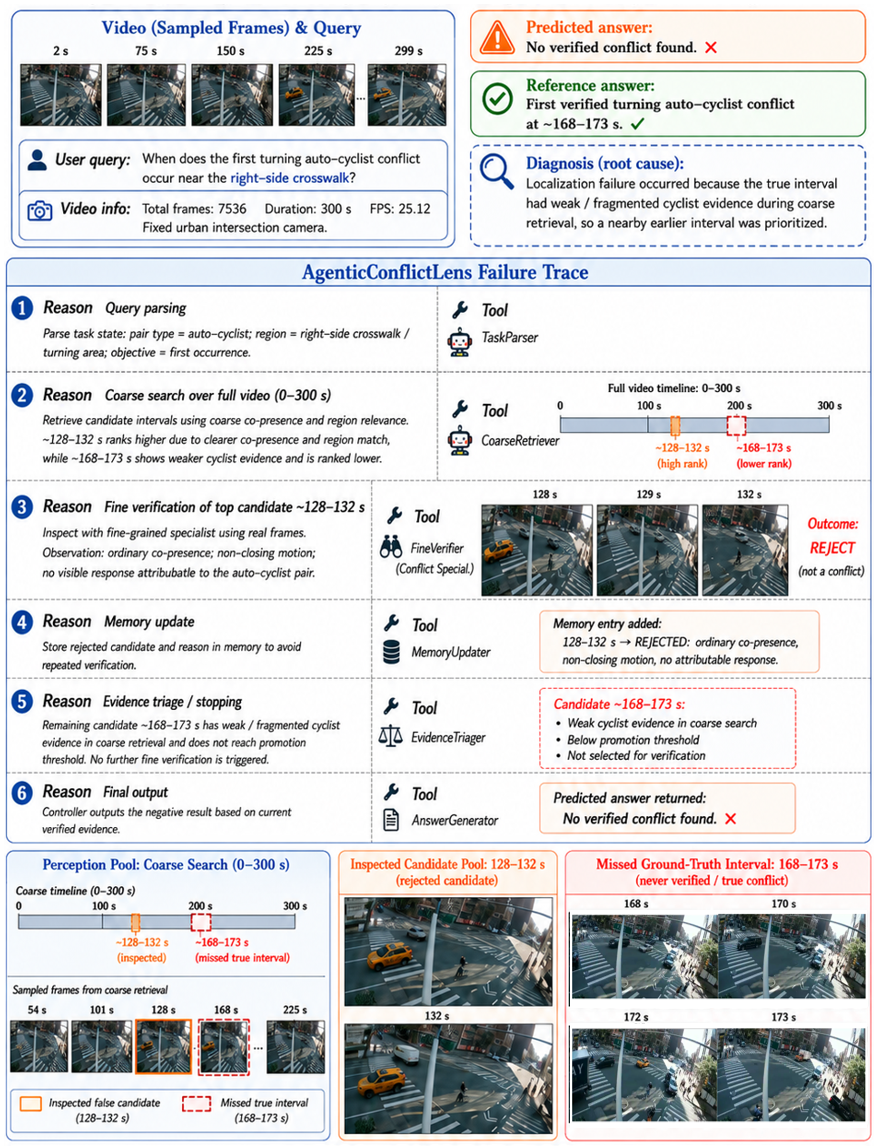}
    \caption{Failure case: a higher-ranked 128--132~s candidate is correctly
    rejected, while the true turning auto--cyclist conflict at approximately
    168--173~s remains below the coarse promotion threshold and is never
    verified.}
    \label{fig:Failure_Case}
\end{figure*}

\paragraph{Failure case.}
Figure~\ref{fig:Failure_Case} shows a turning auto--cyclist conflict at
approximately 168--173~s that is missed because its cyclist evidence is weak
during coarse retrieval. A higher-ranked 128--132~s candidate is inspected and
correctly rejected as ordinary co-presence, but the true interval remains below
the promotion threshold and is never sent to fine verification. This illustrates
a coarse-localization recall failure rather than a verification error.

\subsection{Main Results}
Table~\ref{tab:main_results_detailed} shows that \method{} achieves the best
accuracy--efficiency trade-off on both datasets. It obtains the highest
Exist. F1 and tIoU while inspecting the smallest portion of each 30-minute
query video. On Bellevue, \method{} improves Exist. F1 to 0.84 and tIoU to
0.77 while inspecting only 8.7\% of the video. On NYC DOT, it reaches 0.80
Exist. F1 and 0.73 tIoU while inspecting 11.9\% of the video. SSAM-style
trajectory scoring achieves strong pair accuracy but requires full-video
inspection. The long-video baselines reduce inspection cost, but remain less
accurate and require more frames, calls, and runtime than \method{}.

\begin{table*}[t]
\centering
\small
\caption{Main query-driven participant-level conflict-localization results on
Bellevue (60 annotated hours) and NYC DOT (36 annotated hours). Frames, multimodal calls, and runtime are reported per query for
the 30-minute video; runtime is in minutes. Inspect \% denotes the fraction of
the video sent for detailed inspection.}
\label{tab:main_results_detailed}
\resizebox{\textwidth}{!}{
\begin{tabular}{lccccccc}
\toprule
&
\multicolumn{2}{c}{\textbf{Retrieval / Localization}} &
\multicolumn{1}{c}{\textbf{Pair Quality}} &
\multicolumn{4}{c}{\textbf{Efficiency}} \\
\cmidrule(lr){2-3}
\cmidrule(lr){4-4}
\cmidrule(lr){5-8}
\textbf{Method} &
\textbf{Exist. F1 $\uparrow$} &
\textbf{tIoU $\uparrow$} &
\textbf{Pair Acc. $\uparrow$} &
\textbf{Inspect \% $\downarrow$} &
\textbf{Frames $\downarrow$} &
\textbf{Calls $\downarrow$} &
\textbf{Runtime $\downarrow$} \\
\midrule

\multicolumn{8}{l}{\textbf{Bellevue}} \\

High-risk trajectory intervals~\cite{ssam2008manual}
& 0.74 & 0.66 & \textbf{0.86} & 100.0 & 45216 & 0 & 9.8 \\

Uniform sampled clip conflict search~\cite{zhang2025seeunsafe}
& 0.59 & 0.45 & 0.60 & 62.2 & 28124 & 224 & 6.7 \\

Top retrieved conflict clips~\cite{wang2025videotree}
& 0.62 & 0.57 & 0.64 & 23.5 & 10626 & 96 & 3.8 \\

Memory-selected conflict clips~\cite{yin2026videoarm}
& 0.76 & 0.63 & 0.63 & 23.0 & 10400 & 88 & 3.5 \\

Tool-selected conflict clips~\cite{lin2026videoseek}
& 0.73 & 0.71 & 0.75 & 19.1 & 8636 & 76 & 3.1 \\

\method{}
& \textbf{0.84} & \textbf{0.77} & 0.85
& \textbf{8.7} & \textbf{3934} & \textbf{42} & \textbf{1.6} \\

\midrule
\multicolumn{8}{l}{\textbf{NYC DOT}} \\

High-risk trajectory intervals~\cite{ssam2008manual}
& 0.71 & 0.64 & 0.80 & 100.0 & 45216 & 0 & 10.5 \\

Uniform sampled clip conflict search~\cite{zhang2025seeunsafe}
& 0.60 & 0.44 & 0.61 & 59.7 & 26994 & 215 & 7.9 \\

Top retrieved conflict clips~\cite{wang2025videotree}
& 0.57 & 0.59 & 0.69 & 29.4 & 13294 & 112 & 4.6 \\

Memory-selected conflict clips~\cite{yin2026videoarm}
& 0.70 & 0.61 & 0.62 & 25.9 & 11711 & 104 & 4.2 \\

Tool-selected conflict clips~\cite{lin2026videoseek}
& 0.74 & 0.63 & 0.72 & 22.7 & 10264 & 92 & 3.8 \\

\method{}
& \textbf{0.80} & \textbf{0.73} & \textbf{0.81}
& \textbf{11.9} & \textbf{5381} & \textbf{54} & \textbf{2.1} \\

\bottomrule
\end{tabular}
}
\end{table*}

\subsection{Additional Query Breakdowns}
\label{app:query_extended_results}

Table~\ref{tab:query_difficulty_app} shows that performance decreases as
queries become less constrained. Scoped queries are easiest because they provide
both temporal and participant cues, while open-ended queries are hardest because
the system must search without knowing the conflict time or participant pair.
Across all difficulty levels, \method{} achieves the highest F1 on both
datasets. The gains are largest for actor-conditioned and open-ended queries,
showing that query-conditioned routing, memory, and verification are most useful
when the search space is broad.

\begin{table*}[t]
\centering
\small
\caption{Conflict-existence F1 by query difficulty on Bellevue and NYC DOT.}
\label{tab:query_difficulty_app}
\begin{tabular}{lcccc}
\toprule
\textbf{Method}
& \textbf{Scoped}
& \textbf{Actor-cond.}
& \textbf{Open-ended}
& \textbf{Overall} \\
\midrule

\multicolumn{5}{l}{\textbf{Bellevue}} \\

High-risk trajectory intervals~\cite{ssam2008manual}
& 0.80 & 0.74 & 0.64 & 0.74 \\

Uniform sampled clip conflict search~\cite{zhang2025seeunsafe}
& 0.65 & 0.58 & 0.48 & 0.59 \\

Top retrieved conflict clips~\cite{wang2025videotree}
& 0.68 & 0.61 & 0.51 & 0.62 \\

Memory-selected conflict clips~\cite{yin2026videoarm}
& 0.82 & 0.76 & 0.67 & 0.76 \\

Tool-selected conflict clips~\cite{lin2026videoseek}
& 0.79 & 0.73 & 0.64 & 0.73 \\

\textbf{\method}
& \textbf{0.88} & \textbf{0.84} & \textbf{0.76} & \textbf{0.84} \\

\midrule
\multicolumn{5}{l}{\textbf{NYC DOT}} \\

High-risk trajectory intervals~\cite{ssam2008manual}
& 0.77 & 0.70 & 0.61 & 0.71 \\

Uniform sampled clip conflict search~\cite{zhang2025seeunsafe}
& 0.66 & 0.59 & 0.50 & 0.60 \\

Top retrieved conflict clips~\cite{wang2025videotree}
& 0.63 & 0.56 & 0.47 & 0.57 \\

Memory-selected conflict clips~\cite{yin2026videoarm}
& 0.76 & 0.70 & 0.60 & 0.70 \\

Tool-selected conflict clips~\cite{lin2026videoseek}
& 0.80 & 0.74 & 0.65 & 0.74 \\

\textbf{\method}
& \textbf{0.84} & \textbf{0.80} & \textbf{0.72} & \textbf{0.80} \\

\bottomrule
\end{tabular}
\end{table*}

\subsubsection{Participant and Conflict-Type Breakdowns}
\label{app:participant_conflict_breakdowns}

Table~\ref{tab:participant_results_app} shows that \method{} maintains the
highest F1 across all participant families. Performance is strongest for
auto--pedestrian and auto--auto conflicts, while auto--cyclist cases are harder
because cyclists are smaller, less frequent, and more affected by occlusion and
tracking noise. SSAM-style trajectory scoring remains competitive for
auto--auto interactions, but \method{} performs better for vulnerable road users
and larger-vehicle conflicts by combining participant-aware search with
traffic-specific verification.

\begin{table*}[t]
\centering
\normalsize
\setlength{\tabcolsep}{6pt}
\renewcommand{\arraystretch}{1.15}
\caption{Conflict-existence F1 by participant family on the combined Bellevue
and NYC DOT real-world evaluation.}
\label{tab:participant_results_app}
\begin{tabular}{p{0.50\textwidth}cccc}
\toprule
\textbf{Method} &
\textbf{Auto--Ped. $\uparrow$} &
\textbf{Auto--Cyclist $\uparrow$} &
\textbf{Auto--Auto $\uparrow$} &
\textbf{Bus/Truck $\uparrow$} \\
\midrule

High-risk trajectory intervals~\cite{ssam2008manual}
& 0.72 & 0.66 & 0.75 & 0.70 \\

Uniform sampled clip conflict search~\cite{zhang2025seeunsafe}
& 0.63 & 0.54 & 0.60 & 0.58 \\

Top retrieved conflict clips~\cite{wang2025videotree}
& 0.62 & 0.54 & 0.59 & 0.57 \\

Memory-selected conflict clips~\cite{yin2026videoarm}
& 0.72 & 0.65 & 0.70 & 0.68 \\

Tool-selected conflict clips~\cite{lin2026videoseek}
& 0.76 & 0.68 & 0.74 & 0.71 \\

\midrule
\textbf{\method}
& \textbf{0.83} & \textbf{0.76} & \textbf{0.82} & \textbf{0.79} \\

\bottomrule
\end{tabular}
\end{table*}

\subsection{Additional Experimental Details}
\label{app:complete_ablation}

The main paper reports the principal experimental results. Here, we provide
additional analyses using the same 36 hours of real-world NYC DOT fixed-camera
video from the three intersections included in the main evaluation. No
additional dataset is introduced. Unless otherwise stated, the analyses use
the same videos, queries, backbone, output schema, prompts, thresholds, and
inference configuration as the main evaluation. We examine two complementary
properties of the system on these real-world recordings: the effect of
context-adaptive conflict localization and the spatial organization of verified
conflicts across the three NYC DOT sites.

\subsubsection{Adaptive Conflict Localization: Recovery Checks}
\label{app:adaptive_localization}

We analyze the same 36-hour real-world NYC DOT evaluation data to examine
whether the context-adaptive proximity threshold in
Eq.~\eqref{eq:adaptive_proximity} improves conflict-candidate localization
relative to fixed pair-type thresholds. The comparison uses the same three NYC
DOT intersections, denoted Sites A--C, and changes only the
proximity-thresholding strategy.

\paragraph{Recovery improvement.}
Replacing fixed pair-type thresholds with the context-adaptive rule increases
the number of recovered candidate conflict interactions by approximately 48\%
at Site A, 47\% at Site B, and 49\% at Site C. The similar relative improvement
across the three real-world NYC sites, despite substantial differences in
interaction frequency, indicates that the adaptive rule responds to local
interaction context rather than relying on site-specific threshold tuning.

\begin{figure}[t]
\centering
\includegraphics[width=0.5\linewidth]{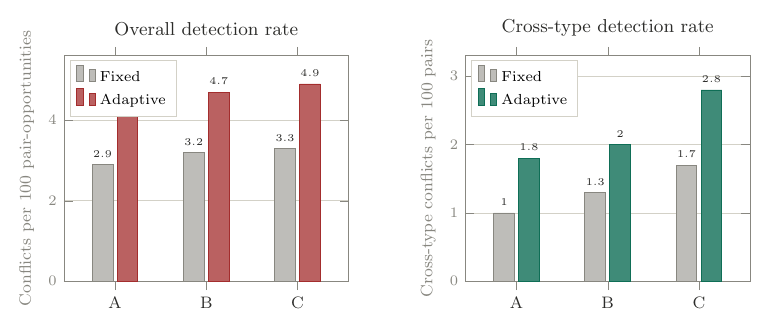}
\caption{Exposure-normalized candidate conflict-interaction recovery under
fixed and context-adaptive proximity thresholds on the same 36-hour
real-world NYC DOT evaluation data at Sites A--C.}
\label{fig:fixed_vs_adaptive_rate}
\end{figure}

\paragraph{Exposure normalization.}
Raw recovery counts depend on traffic exposure because denser intersections
contain more road-user encounters and therefore more opportunities for
interaction. We therefore normalize recovered candidate interactions by
road-user co-presence opportunities. The adaptive improvement remains after
normalization (Fig.~\ref{fig:fixed_vs_adaptive_rate}), indicating that the gain
is not explained solely by differences in traffic volume across the NYC DOT
sites.

\begin{figure}[t]
\centering
\includegraphics[width=0.5\linewidth]{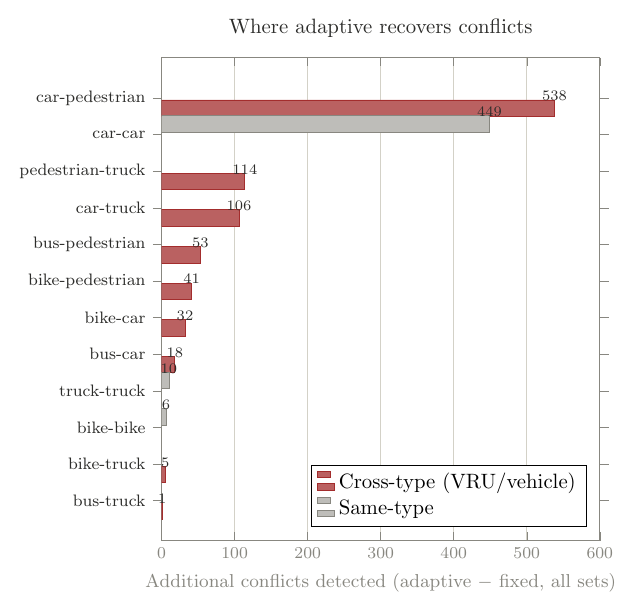}
\caption{Additional candidate conflict interactions recovered by the
context-adaptive proximity rule on the same 36-hour real-world NYC DOT
evaluation data, grouped by participant-pair family.}
\label{fig:recovery_by_pair}
\end{figure}

\paragraph{Composition of recovered interactions.}
The additional candidate interactions recovered by the adaptive rule are not
distributed uniformly across participant classes. Cross-type interactions
involving vehicles and vulnerable road users show the largest relative
improvement (Fig.~\ref{fig:recovery_by_pair}). This pattern suggests that a
single fixed proximity criterion is less effective across heterogeneous
interaction geometries, particularly for participant pairs with substantially
different sizes, speeds, and movement patterns.

\subsubsection{Hotspot Zone Profiles}
\label{app:zone_profiles}

We next analyze the spatial organization of the verified conflict interactions
produced from the same 36-hour real-world NYC DOT evaluation data. Applying the
severity-weighted conflict-density formulation in Eq.~\eqref{eq:density}
identifies twelve hotspot zones across the three intersections: four at Site A,
six at Site B, and two at Site C. The zones are extracted from the
severity-weighted field of verified conflicts using a common normalized density
threshold.

\paragraph{Two zone families.}
Across the three NYC DOT sites, the identified hotspots separate into two broad
behavioral families. \emph{Through-traffic corridors} occur along dominant
vehicle paths and contain predominantly same-type vehicle interactions,
corresponding mainly to merging--following conflicts in the typology of
Appendix~\ref{app:conflict_typology}. \emph{Crossing zones} occur around
crosswalks and turning areas and contain predominantly cross-type interactions,
including vehicle--pedestrian and bicycle--pedestrian encounters, corresponding
mainly to crossing-path and yielding conflicts. These spatial patterns indicate
that the density peaks correspond to distinct traffic-interaction mechanisms
rather than simply regions of high road-user activity.

\begin{figure}[t]
\centering
\includegraphics[width=0.7\linewidth]{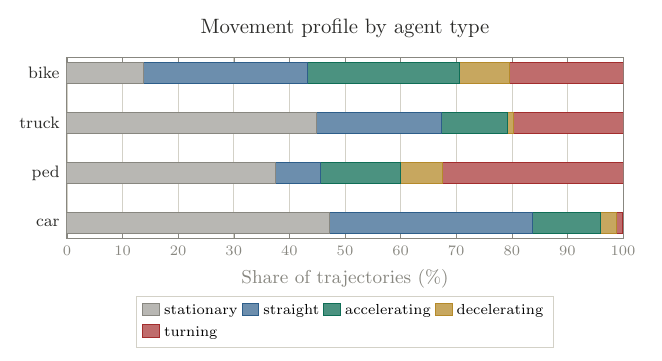}
\caption{Movement profiles by participant type within verified-conflict
hotspot zones identified from the same 36-hour real-world NYC DOT evaluation
recordings.}
\label{fig:movement_by_type}
\end{figure}

\begin{figure*}[t]
\centering
\includegraphics[width=0.3\linewidth]{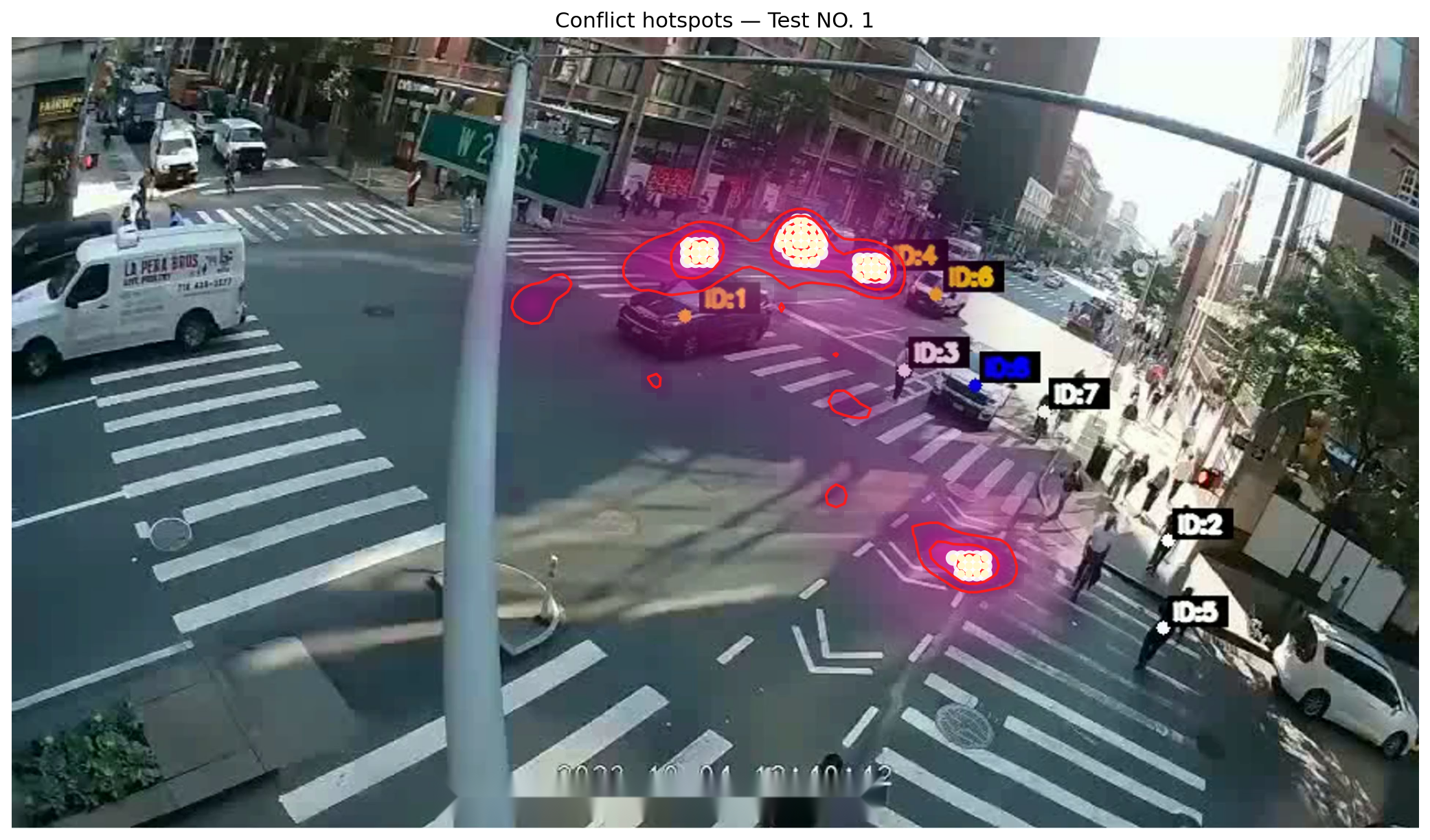}\hfill
\includegraphics[width=0.3\linewidth]{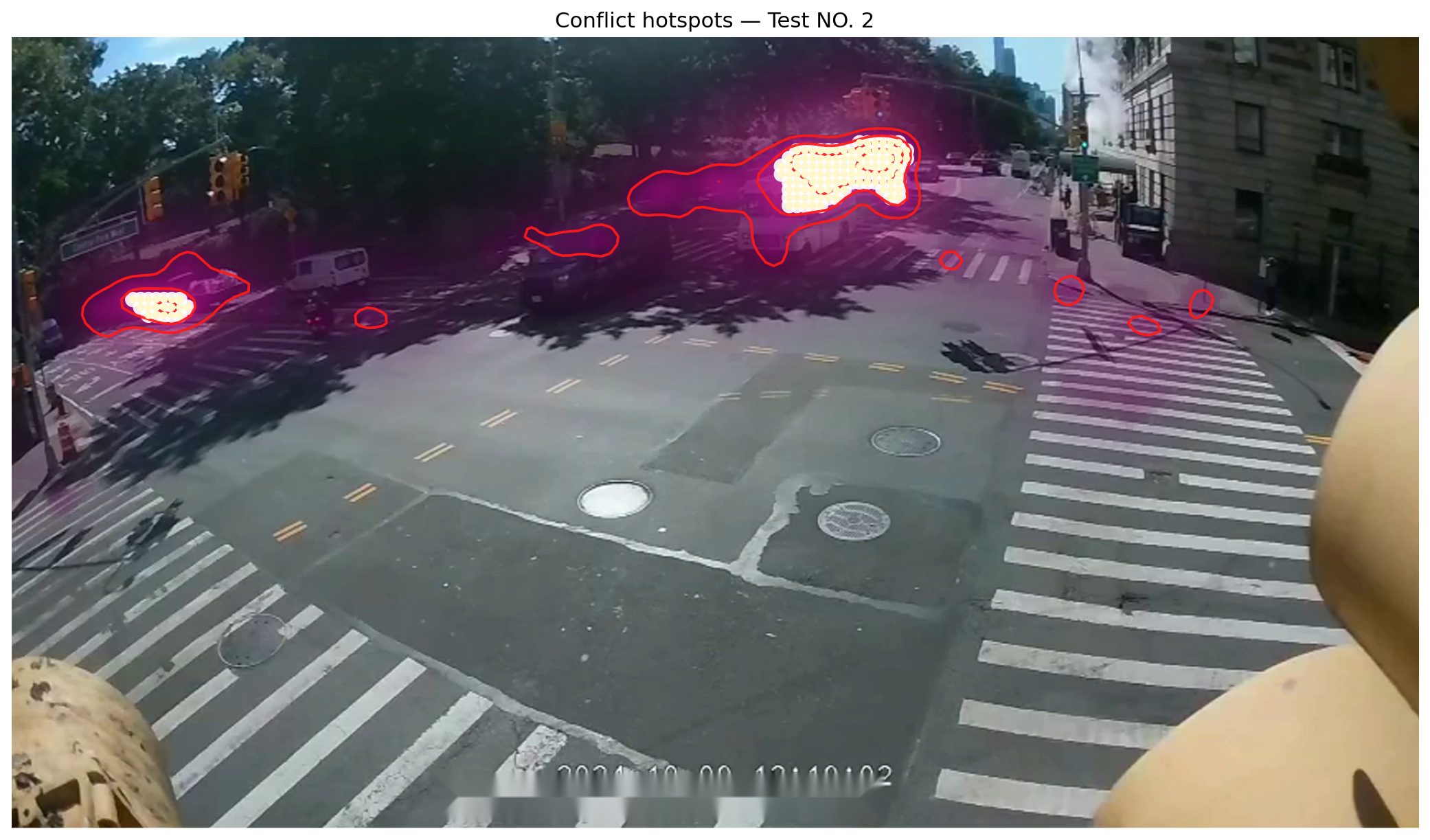}\hfill
\includegraphics[width=0.3\linewidth]{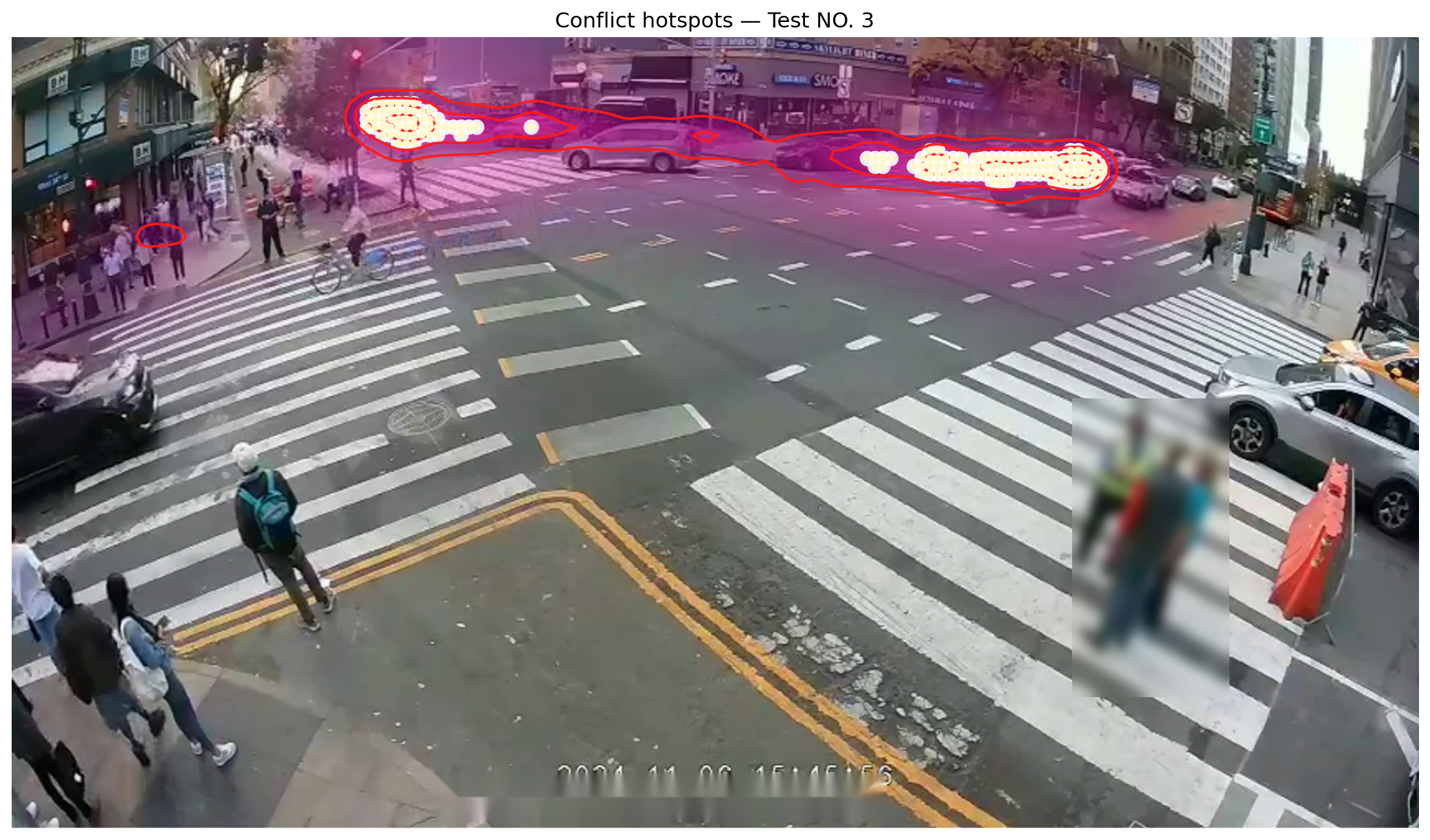}
\caption{Severity-weighted verified-conflict density $\tilde{D}$
(Eq.~\eqref{eq:density}) over the real-world NYC DOT camera views at
Sites A--C. Iso-density contours identify four, six, and two hotspot zones,
respectively, from the same 36-hour NYC DOT evaluation data.}
\label{fig:hotspot_maps}
\end{figure*}

\paragraph{Movement profiles.}
Per-participant movement statistics within the identified hotspot zones
(Fig.~\ref{fig:movement_by_type}) provide additional kinematic support for
this separation. Vehicle-dominated through-traffic corridors exhibit higher
speeds and longer straight-line trajectories, whereas crossing zones contain
lower-speed motion and larger heading changes. Thus, the two zone families are
also distinguishable from their underlying movement characteristics,
independently of the conflict-type labels used to describe them.

\paragraph{Hotspot zone extraction.}
Hotspot zones are defined as connected components of the super-level set
\[
\Omega_\gamma =
\{\mathbf{x} : \tilde{D}(\mathbf{x}) \geq \gamma\},
\]
using a common normalized density threshold $\gamma$ and discarding components
below a minimum area. Each retained component is summarized by its spatial
extent and peak-density location
($\arg\max \tilde{D}$ within the component).
Figure~\ref{fig:hotspot_maps} renders the normalized verified-conflict density
field directly over the three real-world NYC DOT camera views, making the
resulting spatial structure visually interpretable. Because $\tilde{D}$ is
normalized independently for each intersection, the same thresholding
procedure is applied consistently across Sites A--C. The resulting zones
provide a compact spatial summary of the verified conflict interactions
observed in the real-world NYC DOT evaluation data.






\end{document}